\documentclass{elsarticle}
\usepackage{setspace}
\usepackage{url}
\usepackage{algorithm}
\usepackage{algpseudocode}
\usepackage[utf8]{inputenc}
\usepackage{graphicx}
\usepackage{epsfig}
\usepackage{booktabs}
\usepackage{amsmath}
\usepackage{mathrsfs}
\usepackage{amsfonts}
\usepackage{lscape}
\usepackage{makecell}
\usepackage{longtable}
\usepackage{subfigure}
\usepackage{amssymb}
\usepackage{natbib}
\newcommand{\BORRAR}[1]{}

\DeclareMathOperator{\argmin}{\textsf{argmin}}
\DeclareMathOperator{\argmax}{\textsf{argmax}}
\begin{document}

\begin{frontmatter}

\title{Improving classification performance by feature space transformations and model selection
}

\author[infotec,umsnh]{Jose Ortiz-Bejar}
\ead{jortiz@umich.mx}
\author[infotec,conacyt]{Eric S. Tellez}
\ead{eric.tellez@infotec.mx}
\author[infotec,conacyt]{Mario Graff}
\ead{mario.graff@infotec.mx}

\address[infotec]{INFOTEC Centro de Investigaci\'on e Innovaci\'on en Tecnolog\'ias 
de la Informaci\'on y Comunicaci\'on, Aguascalientes, M\'exico}
\address[umsnh]{Divisi\'on de Estudios de Postgrado \\Facultad de
  Ingenier\'ia El\'ectrica \\Universidad Michoacana de San Nicol\'as de
  Hidalgo, Michoac\'an, M\'exico}
\address[conacyt]{CONACyT Consejo Nacional de Ciencia y Tecnolog\'ia,
Direcci\'on de C\'atedras, Ciudad de M\'exico, M\'exico}

\begin{abstract}
Improving the performance of classifiers is the realm of feature mapping, prototype selection, and kernel function transformations; these techniques aim for reducing the complexity, and also, improving the accuracy of models. In particular, our objective is to combine them to transform data's shape into another more convenient distribution; such that some simple algorithms, such as Na\"ive Bayes or k-Nearest Neighbors, can produce competitive classifiers.
In this paper, we introduce a family of classifiers based on feature mapping and kernel functions, orchestrated by a model selection scheme that excels in performance. We provide an extensive experimental comparison of our methods with sixteen popular classifiers on more than thirty benchmarks supporting our claims.  In addition to their competitive performance, our statistical tests also found that our methods are different among them, supporting our claim of a compelling family of classifiers.

\end{abstract}
\end{frontmatter}

\section{Introduction}
\label{sec:introduction}


The complexity of classification problems has been continuously increasing due to the availability of more and diverse sources of information. The more detailed characterization of the data along with the amount of information, impose challenges for information processing techniques to remain competitive.

An automatic classifier is a model that can predict the class of a given object based on a learning process performed on a training set, i.e., a set of examples $X=\{x_1,\cdots,x_n\}$ and its associated labels (linked to the valid classes) $y=\{y_1,\cdots,y_n\}$. Note that the training set is part of a universe of valid objects $X \subset U$; also, $U$ contain all inputs for the model. Automatic classifiers are a fundamental part of more complex tasks in many fields of science and industrial processes. While there exists plenty of algorithms, to create classifiers, this manuscript focus on kernel methods and its practical exponents. Kernel methods are a family of algorithms that need a particular function definition over pairs of objects in $U$ to work — the rest of the document detail more about the kernel functions and methods.

Prototype selection approaches aim to reduce the number of examples taking a subset of elements that describe the properties of the complete database, or by generating elements that summarize the available information, see \cite{triguero2012}. Both selection and generation of prototypes can be seen as sampling methods that retrieve points from a distribution.
On the other hand, techniques for feature selection deal with the problem of selecting and generating relevant features that correctly describe all element's classes of the database, see \cite{tang2014feature}. Both approaches can be understood as the selection of rows (elements) or columns (features) in a matrix of samples. However, even though creating lower-dimensional data implies less memory and computing resources; the elimination of high correlated features and elements does not always result in better classification accuracy. At this point, feature mapping and kernel-based methods come to play. Notice that both methods are not oriented to reduce data's dimensionality, but its objective is to project data into a new feature space where the problem can be easily classified.

\begin{figure}[!ht]
\centering
\subfigure[An almost linear separable set of points.]{
    \label{fig/linearsep}
    \centering\includegraphics[width=0.31\textwidth, trim={1.5cm 1.2cm 1.2cm 1.5cm}, clip]{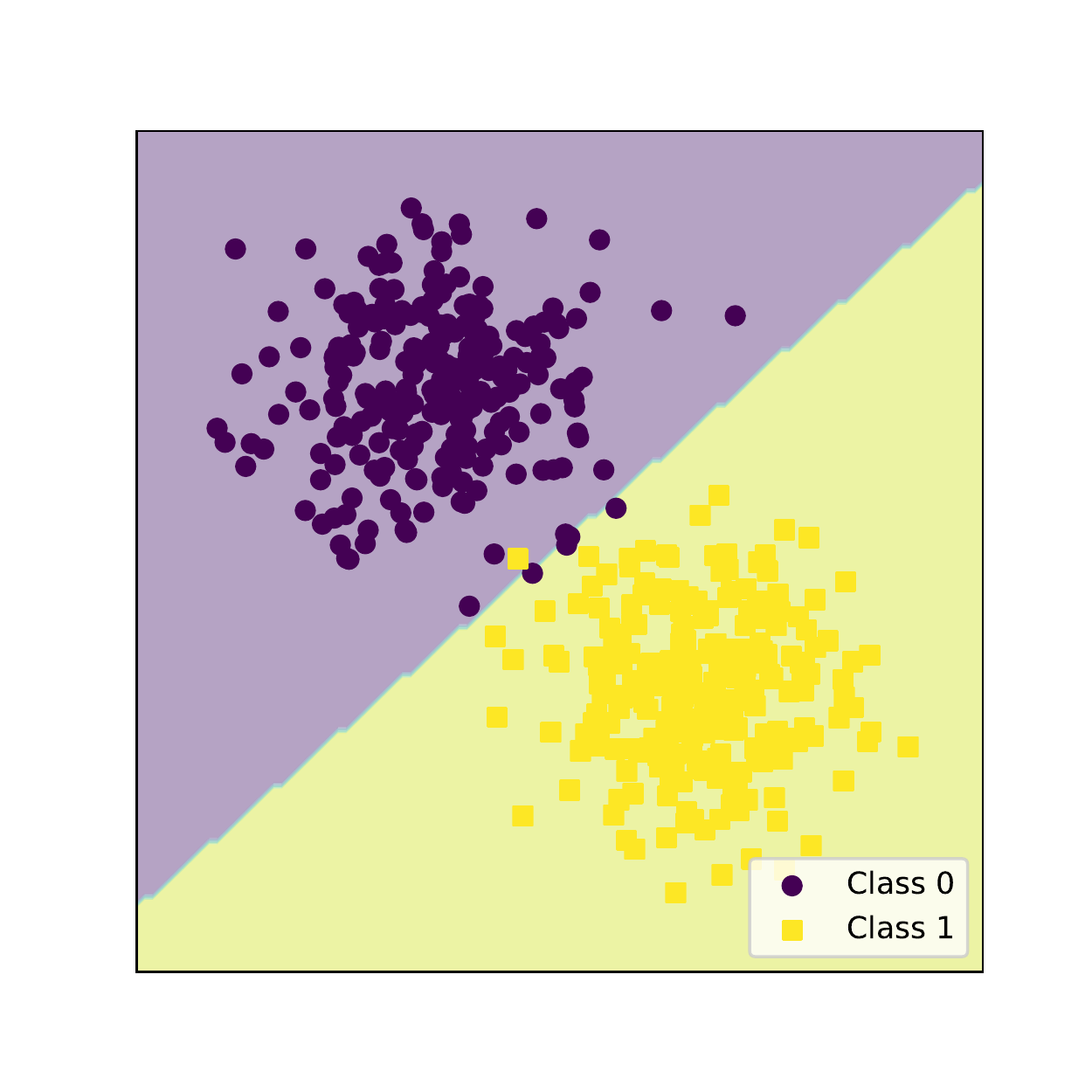}
}\hfill\subfigure[A non-linearly separable dataset.]{
    \label{fig/nonlinearsep}
    \centering\includegraphics[width=0.31\textwidth, trim={1.5cm 1.2cm 1.2cm 1.5cm}, clip]{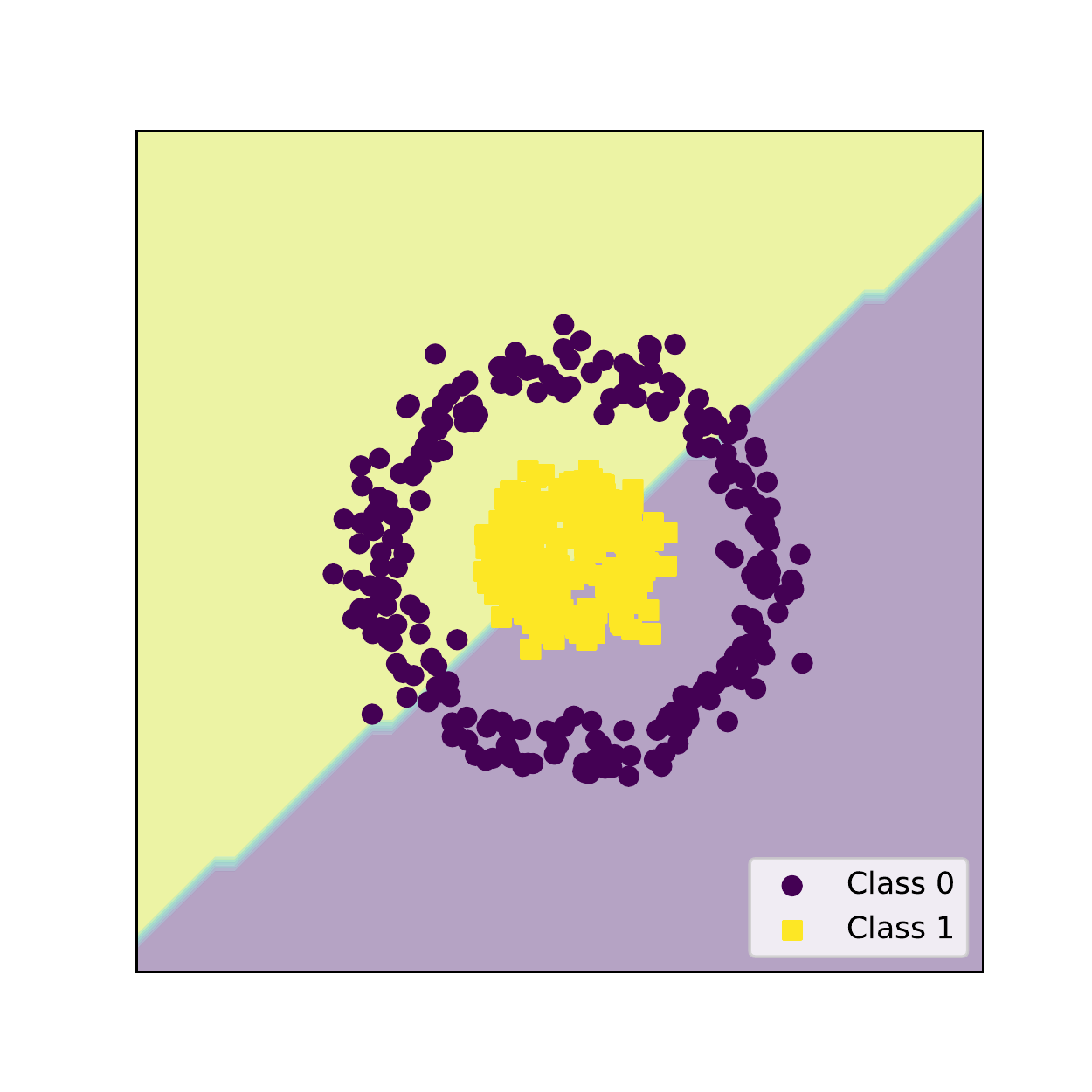}
}\hfill\subfigure[A linear separable dataset after transforming points in Fig.~\ref{fig/nonlinearsep}.]{
    \label{fig/projection}
    \centering\includegraphics[width=0.31\textwidth, trim={3.5cm 1.5cm 2.8cm 2.6cm
    }, clip]{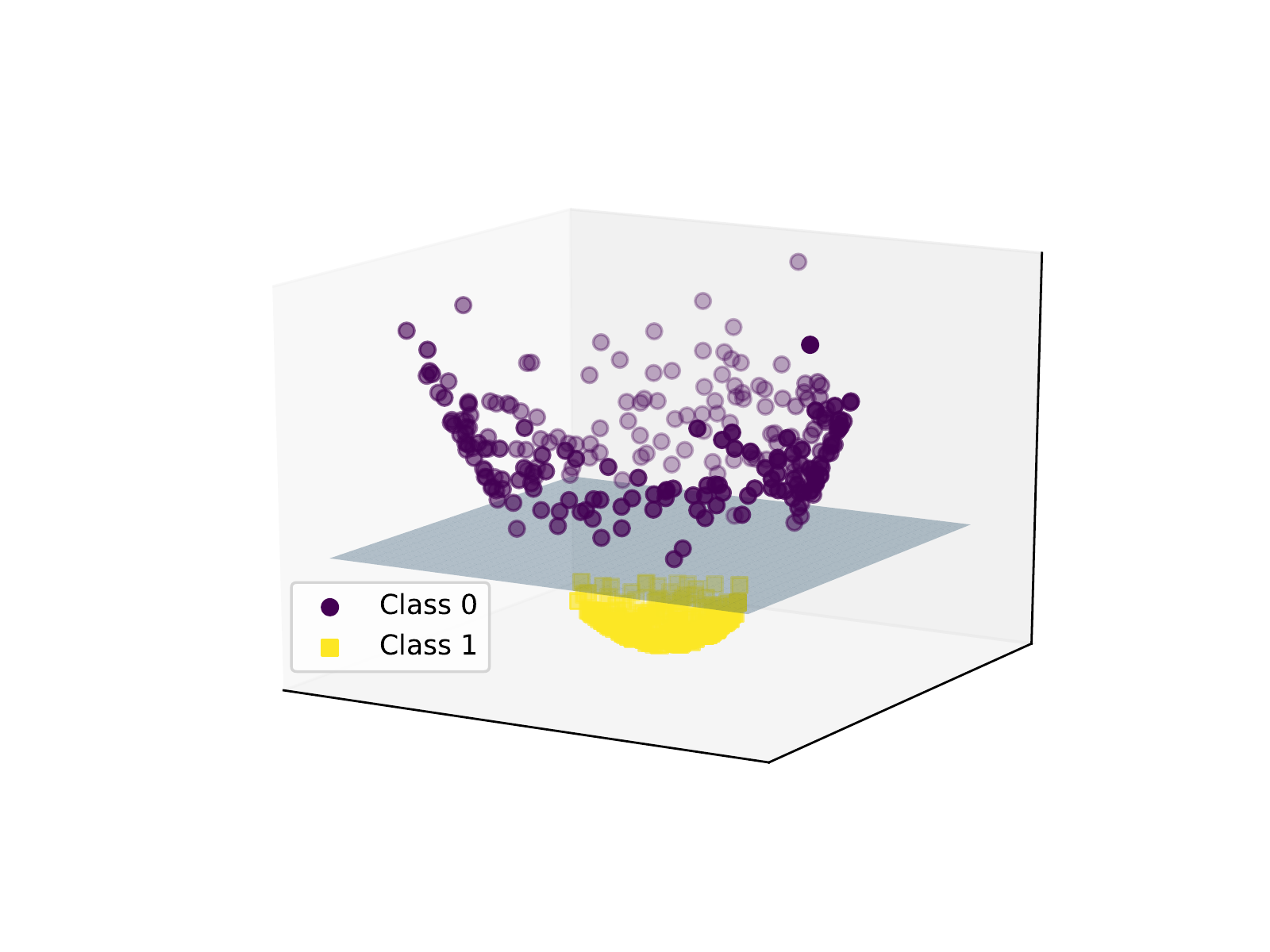}
}
\caption{Example of linear and non-linear separable binary classification problems.}
\label{fig/sep-examples}
\end{figure}

It is worth to notice that kernel-based methods have been widely used in industrial-strength applications due to its excellent performance. For instance, consider a linear classifier, capable of performing an accurate classification if the data is linearly separable like that illustrated in Figure~\ref{fig/linearsep}, the classifier only needs to distinguish between items on the left and right of the dividing hyperplane. On the other hand, if the dataset cannot be linearly separated, like Figure~\ref{fig/nonlinearsep}, then a linear classifier fails in finding a good performing model for the data. Instead of discarding the linear classifier, an option is to transform the shape of the dataset using a kernel function (maybe a non-linear one) such that a hyperplane can perform a proper division of the classes. A way to do this is projecting the input dataset into a lower or higher dimension space, for example, Figure \ref{fig/projection} shows a projection of data in Figure~\ref{fig/nonlinearsep} in space where a linear separation is possible.

Commonly, kernel methods tend to increase the data's dimensionality; fortunately, this situation is efficiently handled since kernel methods can be expressed in terms of a dot product whenever the kernel function follows Mercer's condition. This technique is the so-called {\em kernel trick}~\cite{muandet2017kernel,murphy2012machine}.
The trick has the advantage that works on the original space, so it is not necessary to carry out the explicit mapping.

A simple version of a kernel method classifier can be exemplified by using $k$-Nearest neighbor ($k$NN) method, this is a non-parametric algorithm that uses all known observations of the training set composed by the set of points $\mathbb X = \{x_i\}$, and its associated set of labels $y = \{ y_i \}$, to predict outcomes based on a similarity function; $k$NN is flexible enough to work with both similarity and dissimilarity functions. When $k=1$ the method works as follows: Given a vector $u$, its nearest neighbor $x_i \in X$ is located, then the label $\theta_i$ is associated with $u$. That is, for a dot product similarity, the most similar object is computed as $\argmax_{1 \leq i \leq |X|} \langle u, x_i \rangle$ or $\argmin_{1 \leq i \leq |X|} d(u, x_i)$ in the case of a distance function, i.e., $d: \mathbb R \times \mathbb R \rightarrow \mathbb R^+$ (a two argument function maps to a positive real).  

Some times it is better to use more neighbors to predict; when $k > 1$ the prediction can be computed as the most popular label among $k$ nearest neighbors. The similarity between $u$ and some neighbor $x$ can be used to weight the possibility of some label.

\begin{figure}[!htb]
\centering
\subfigure[Euclidean]{
    \label{fig:knnlinear}
    \centering\includegraphics[width=0.31\textwidth, trim=1.2cm 1.2cm 1.2cm 1.2cm,clip]{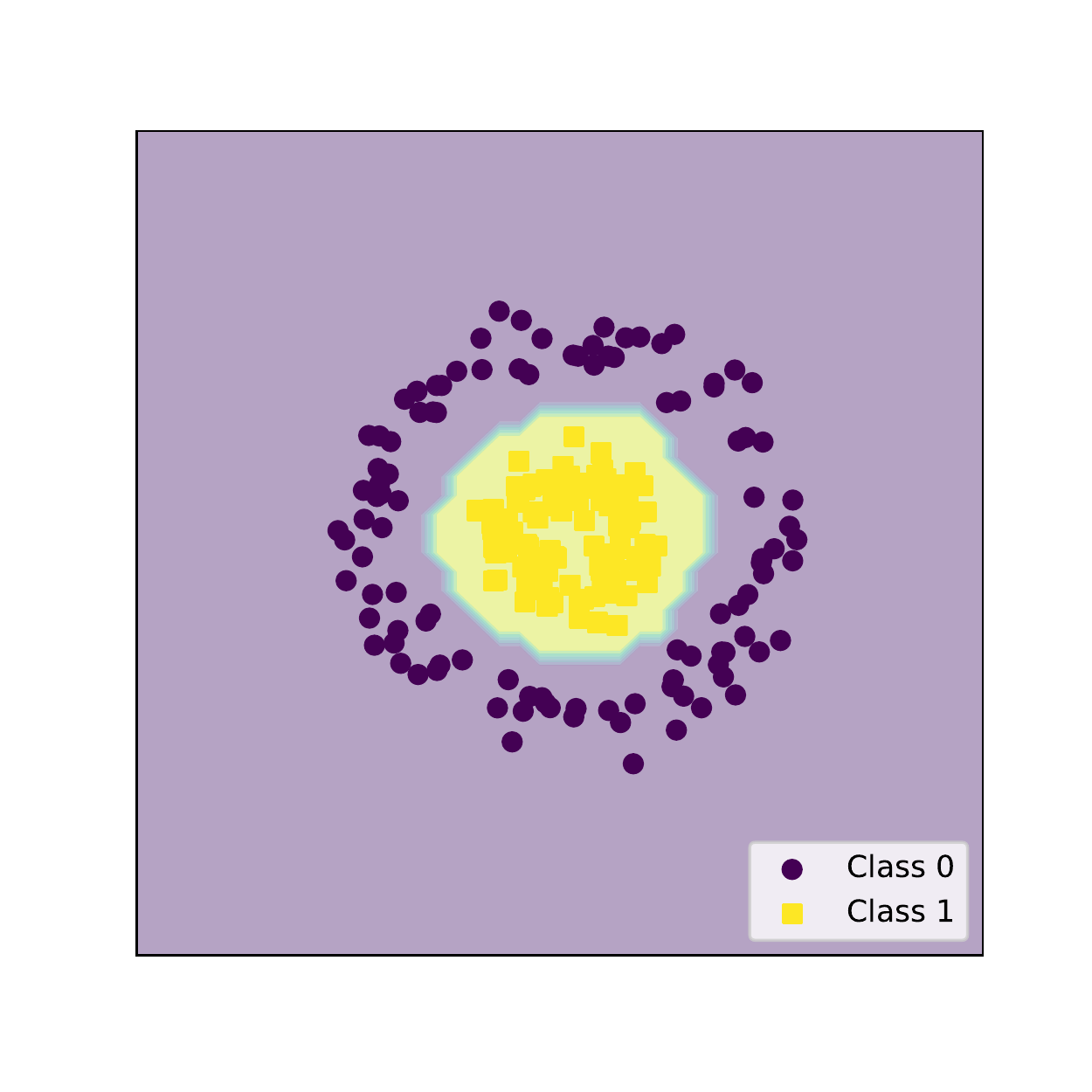}
}\hfill\subfigure[Angle]{
    \label{fig:knncosine}
    \centering\includegraphics[width=0.31\textwidth, trim=1.2cm 1.2cm 1.2cm 1.2cm,clip]{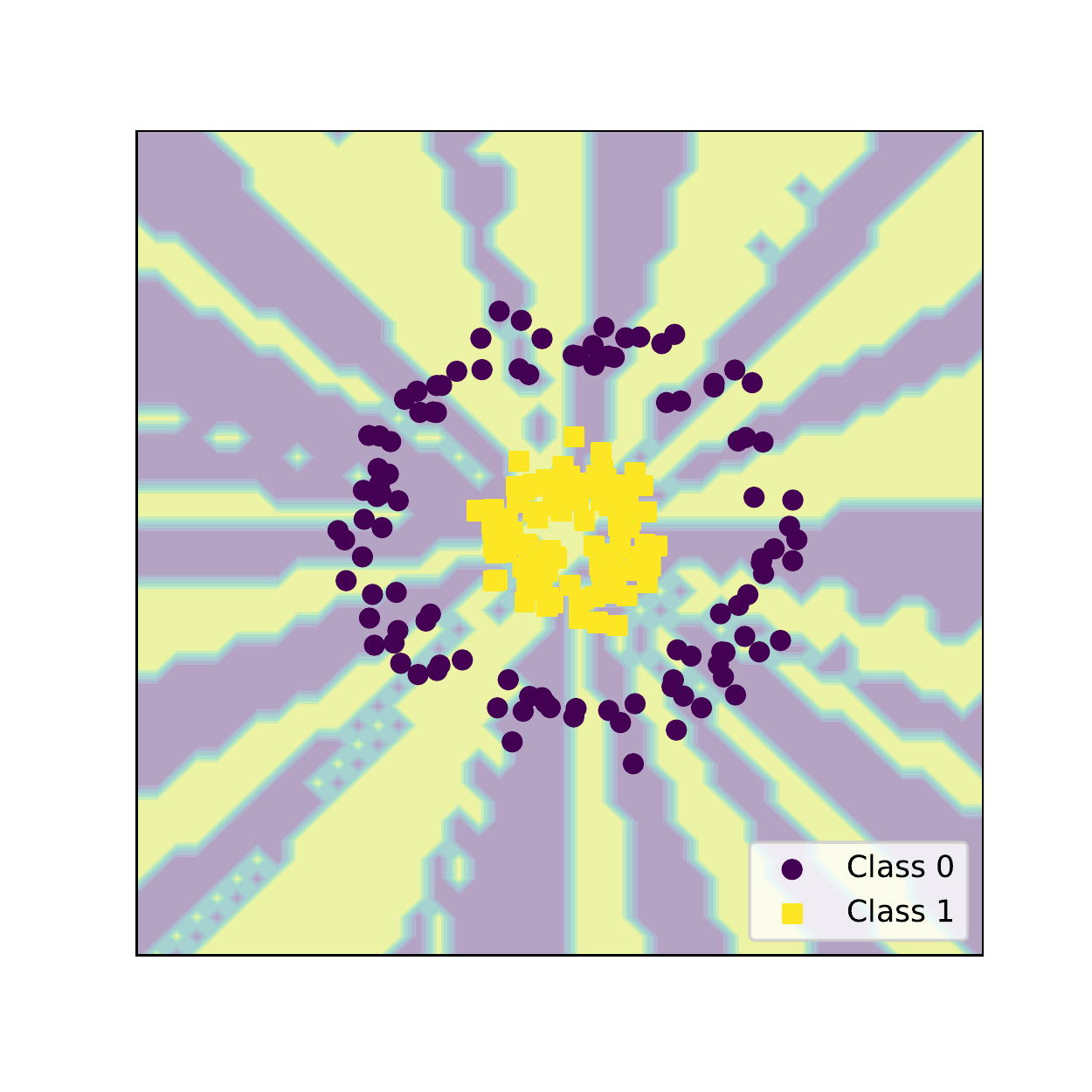}
}\hfill\subfigure[Jensen-Shannon]{
    \label{fig:knnjensen}
    \centering\includegraphics[width=0.31\textwidth,trim=1.2cm 1.2cm 1.2cm 1.2cm,clip]{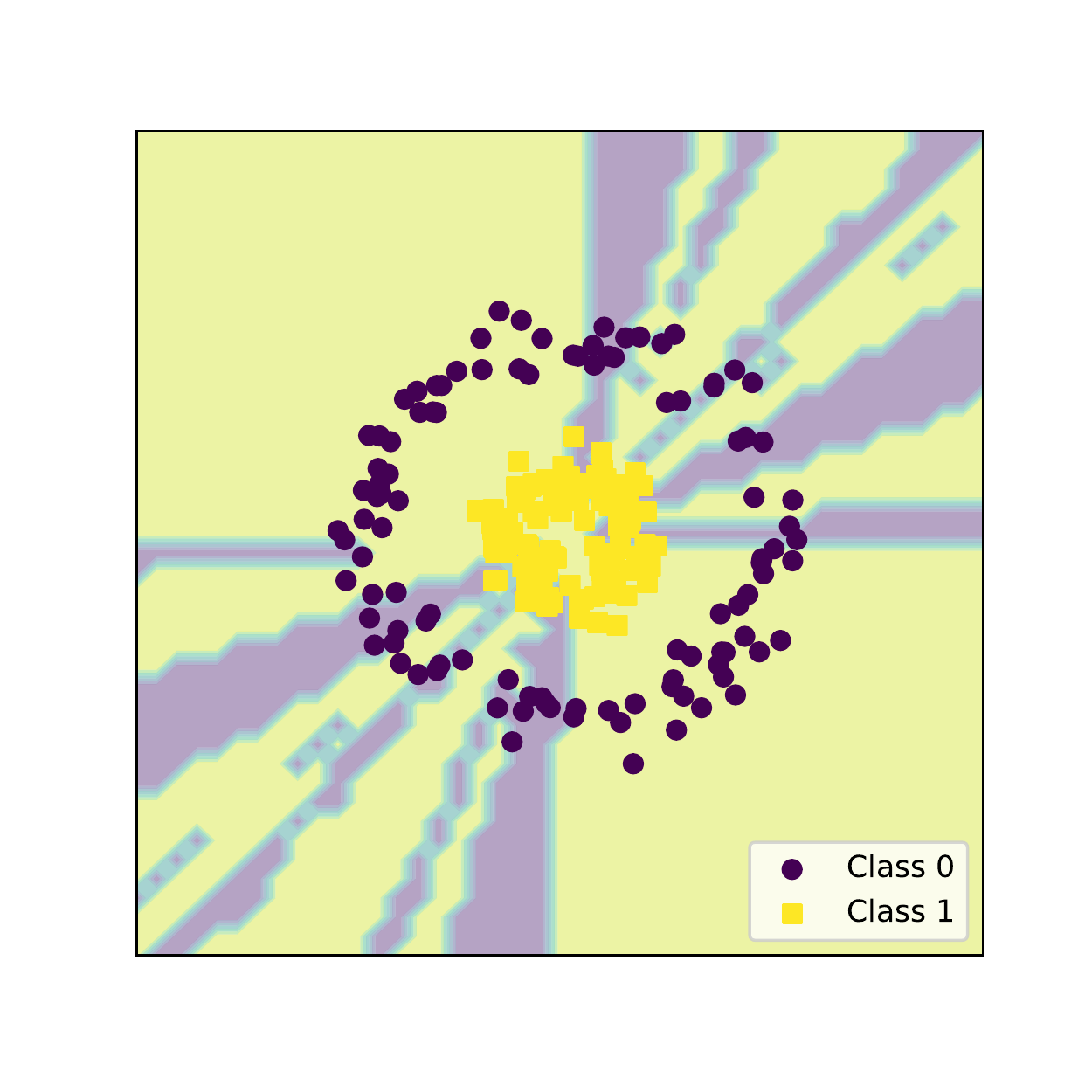}
}
\caption{Decision boundaries for $k$NN with different distance functions; yellow points are predicted with light-yellow regions and purple with light-purple regions.}
\label{fig:knnkernel}
\end{figure}

Figure \ref{fig:knnkernel} shows the learned decision boundary for a 2-dimensional binary class toy-dataset of two concentric circles. The classifier uses $k=5$, uniform weighting,  a linear kernel, and three different distance functions:  Euclidean distance, the angle between vectors as distance, and the Jensen-Shannon divergence as a distance metric, see~\cite{endres2003new}. As the figure illustrates, $k$NN has a non-linearity nature, even with a linear kernel. For instance, the Euclidean distance achieves better performance in this example.

Conveniently, the $k$NN approach does not reside on the sense of global separation, like Support Vector Machines (SVM)~\cite{Joachims/99a}, but in a local-sense of class separation. This strategy helps in solving different kinds of problems without the need for sophisticated kernel functions. 



\subsection{Contribution}
In this manuscript, we propose a new family of classifiers based on a pipeline that includes a feature mapping and the use of a simple classifier, such as k-Nearest Neighbors and Na\"ive Bayes. The feature mapping is made by firstly perform a prototype selection using several strategies that capture different properties of the training dataset; these prototypes are then used to produce a new feature space with the help of several kernel functions. This procedure was inspired in the works related to Nystr\"om method. The prototype selection is based on the application of different sampling and clustering methods, like a random selection, K-Means, Density-nets, and the Furthest First Transversal algorithm; we also consider several kernel functions to produce both linear and non-linear transformations. These components are included in a model selection scheme to decide a competitive combination of them through a pipeline. We also found that ensembling several instances of these classification pipelines we can improve performance at the cost of higher computational resources.


\subsection{Roadmap}
This manuscript is organized as follows. The current section introduces our contribution, the fundamentals of kernel-based methods, the kernel trick, and the $k$ nearest neighbors classifier. Section~\ref{sec:related-work} reviews the related work. 
Our approach is described in Section \ref{sec:okcc}. Section \ref{sec:results}, the results, and a comparison with other states of the art classifiers are carried out. Finally, some conclusions and future research directions are given in Section~\ref{sec:conclusions}.



\section{Related Work}
\label{sec:related-work}


Kernel methods have been proved to be useful to improve many machine learning approaches, some examples are: fisher discriminant \cite{mika1999fisher,liu2004improving}, Support Vectors Machines \cite{vapnik2013nature,guo2014feature,suykens1999least},  manifold learning \cite{coifman2006diffusion,berry2016local,belkin2003laplacian}, more recently with Non-parametric density estimation based classifier \cite{sheikhpour2017kernelized} and even though as Neural Networks activation function \cite{scardapane2019kafnets}. These methods depend on a kernel function, many times represented as a similarity/distance matrix $K$, which is given by the kernelized dot product of each element in the dataset. Since computing $K$ for a large matrix is expensive, in terms of computing time and memory, much research has been done to avoid full computation and storage of $K$. A common approach to overcome the cost associated to $K$ is to create the approximation matrix $\tilde{K}$, where the Nystr\"om method altogether with different sampling procedures for column selection is a widely used strategy ~\cite{drineas2005nystrom,homrighausen2016nystrom,drineas2018lectures}. Nystr\"om approximation stands that it is possible to apply any kernel method over $\tilde{K}$ with little impact on the quality of the result \cite{williams2001using}. This approach is mainly oriented to get a fixed number of eigenvectors efficiently \cite{drineas2018lectures}; these methods are often linked to Singular Value Decomposition (SVD), Principal Component Analysis, and QR factorization. In the literature, there are many sampling methods inspired by Nyst\"orm approximation, which ranges from random sampling \cite{williams2001using}, distributional \cite{drineas2005nystrom}, Ridge Leverage Score~\cite{musco2017recursive}, among others. Kumar et al.~\cite{kumar2012sampling} provides a comprehensive survey into the field. The state-of-the-art sampling strategy for Nyst\"orm method is described at \cite{zhang2009density,zhang2010clustered}, it is based on the usage of K-Means algorithm for computing column's centroids; centroids are then used as references to map original data to $\tilde{K}$. Further analysis and improvements of the K-Means method are reported at \cite{he2018kernel,wang2019scalable} where authors determined the K-Means is optimal regarding $||K-\tilde{K}||$ as error function.

Coifman and Lafon~\cite{coifman2006diffusion} avoid using all columns by applying the incomplete Cholesky decomposition to select the $k$ most relevant columns; then, these columns are used to create a low-rank matrix $\tilde{K}$ where Kernelized Principal Components Analysis (KPCA) was performed. Similarly, Baudat and Anouar~\cite{baudat2003feature} state the column selection as an optimization problem. The training set, $X$, is expressed as a linear combination of the subset of columns $\hat{X} \subset X$; $\hat{X}$ is created by using columns that minimize the normalized Euclidean distance between each column vector $x_i$ and its projection by using $R$ and a weights vector $w$, where the number of features vectors, as well as their weights, must be determined. Lui and Zio \cite{liu2016feature} describe a further modification to apply this approach to regression problems; the authors formulate the problem as a least-squared error optimization problem with equality of constraints. The hyper-parameters are optimized with Grid Search~\cite{BurkeKendall2014}, that is, evaluating a grid of pre-established parameter values.

On the other hand, prototype selection algorithms tackle the problem of large datasets by reducing the number of items in a dataset. Further, a carefully selected set of prototypes can also improve the overall performance of a classification task. Perhaps the most basic representative of this family of methods is the Nearest Centroid (NC) classifier. This method computes one centroid per class, i.e., through the geometric dimension mean. The classification procedure is pretty simple, that is, the class of a non-labeled exemplar is computed as the class of its nearest centroid~\cite{john2010elements}. Despite its simplicity, NC is powerful enough to be used in many applications, mainly for text categorization, due to its simplicity, efficiency, and its proved performance on text classification~\cite{cardoso2007}. Nonetheless, prototype selection algorithms many times, target k-Nearest Neighbor algorithms instead of NC.

We can found multiple methods for prototype generation in literature. In~\cite{liu2017new}, an optimization method, inspired by the gravitational model, is used to determine a weighted mass factor for each prototype; this method is especially useful for imbalanced data sets. In \cite{li2016kernel}, the initial centroids are optimized by minimizing the hypothesis margin under the structural risk minimization principle; and finally, the kernel method is used to deal with linear inseparability in the original feature space. All the above techniques are applied mainly to text classification, to the best of our knowledge, there are not centroid based classifiers that tackle general classification problems. 

A generalization of the center-based classifier can be seen as cluster-based classifier where more than one centroid per class is chosen, many examples of this kind of approaches come from prototype selection. According with \cite{triguero2012}, the most widely used algorithms of this type that present the best results in their respective publications are: Self-generating prototypes (SGP) \cite{fayed2007}, Reduced Space Partition (RSP) \cite{sanchez2004} and Pairwise Opposite Class-nearest Neighbor (POC-NN) \cite{raicharoen2005}. All methods mentioned above aim to split the dataset in a set of clusters where elements at each cluster are homogeneous, i.e., all of them belong to the same class, by always using class boundaries elements; therefore, the main difference among these methods is how elements are selected. For instance, SGP uses hyperplanes and singular value decomposition, while RSP selects the furthest elements at each non-homogeneous cluster. In the case of POC-NN, the cluster division is led by POC-NN prototypes which are used as locations for setting separating hyperplanes. De Brabanter et al.~\cite{de2010optimized} propose an alternative algorithm for prototype selection based on the approximation of the kernel matrix; authors select a subset of the elements that maximize the quadratic R\'enyi entropy criterion. This algorithm is oriented to optimize fixed-size least squares support vector machines, introduced in \cite{suykens1999least}.


Our approach is similar to prototype selection since a set of references are generated; these references can be either centroids or centers. In our context, centroids are prototypes generated by summarizing elements at each group; on the other hand, centers are prototypes being part of the training dataset. The precise strategy depends on the problem. Nonetheless, the main difference with other state-of-the-art approaches is that references are selected in an unsupervised manner. Note that this strategy may also suffer from different problems, like the lack of local support for creating a decision function. In this article, we use different strategies to tackle this problem based on clustering and the k-center problem \cite{gonzalez1985}. In this road, a clustering algorithm based on compactness of the space is required, since the proposed methodology borrows the idea of clustering based prototype's selection and kernel methods to generate a new feature space (not necessarily linear separable) where simple classifiers like $k$-Nearest Neighbors ($k$NN) and Na\"ive Bayes (NB) might achieve competitive performance. For our scheme, we use $k$NN classifier since it is based on similarity (and dissimilarity) measurements, and this is the core of kernel methods. Additionally, we include NB because the new feature space may promote the independence of variables. Nevertheless, there is no restriction to use other classifiers instead of the $k$NN and NB.

Moreover, the set of references are no used to train a classifier directly; instead, they are used to project original data to a new feature space with the help of a kernel function. Our proposal is highly related to Nystr\"om's K-Means sampling, and the central differences are that Nystr\"om methods are usually applied to approximate the top $k$ eigenvalues to perform features reduction (since it is column-wise), while our method aims to locate border references for a classification task, as our method is row-wise. Moreover, most of Nystr\"om methods sample directly from $K$, whereas our approach samples over the original data. Additionally, our method recognizes that many parameters are highly dependent on the classification task, so we state our problem as a model selection problem that finds a competitive classifier among a broad set of possible ones. The following section is dedicated to detail our contribution.

\section{A model selection approach for kernel methods}
\label{sec:okcc}
For our purposes, a classifier is a function $h : \mathbb{R}^d \rightarrow \mathcal{L}$; that is, $h$ maps a real-valued $d$-dimensional vector to a member of $\mathcal{L}$, which is a set of categorical values. In particular, $h$ is an item among $\mathcal{H}$, the infinite set containing all possible functions with the $h$'s signature. Given a classification task $({X}, y, \textsf{err})$, the idea is to find a function $h \in \mathcal{H}$ that reaches an acceptable error ratio under a cross-validation scheme. The selection of such $h$ is known as {\em training step}. Testing set ${X',y'}$ is used to validate $h$'s performance; the testing set is no available during the training step.

In this context, ${X}$ and ${X'}$ are subsets of $\mathbb{R}^d$, of size $n$ and $m$, respectively; on the other hand, $y$ and $y'$ are subsets of $\mathcal{L}^n$ and $\mathcal{L}^{m}$, respectively. Finally, the function $\textsf{err}: \mathcal{L}^{m} \times \mathcal{L}^{m} \rightarrow \mathbb{R}^+$ computes the fitness between its arguments. Therefore, the training step is the process of finding $h$ such that the function $\textsf{err}(y, h({X}))$ is minimized. For our experiments we measure the Balanced Error Rate (BER) which is defined as the average proportion error, per-class, as follows: $$\text{BER} = \frac{1}{|\mathcal{L}|} \sum_{c \in \mathcal{L}} \frac{\text{false positives}_{c} + \text{false negatives}_{c}}{\text{\#samples}_{c}}.$$ Since smaller BER values are better than higher ones, we are interested in $h$ functions such as that $\textsf{err}(y, h({X}))$ is minimal.




Our method searches for the parameters $(f, R, C)$,  where $f$ correspond to the kernel, $R$ is the set of references, and $C$ is the classifier, that can prove its competitive performance for the given classification task $({X}, y, \textsf{err})$. These parameters, jointly, define a {\em configuration}; the set of all possible configurations is named a {\em configuration space}.  The tuple $(f, R, C)$ is a meta-specification of the configuration space to search for, which in turn represent functions in $\mathcal H$. In particular, the selection of these three parts can be tightly linked, and, consequently, these must be selected jointly. Since we use stochastic algorithms to explore the configuration space, it is desirable to produce high-quality predictors of the performance of each configuration; we choose to embed a cross-validation scheme into the \textsf{err} function.

In summary, we can explain our approach as the following pipeline. Given a training dataset, we select a subset of it based on a distance function and a sampling method (we call this subset as references); the sample is then used, along with a kernel function, to map the original space into a new {\em kernelized} space. Afterward, an internal classifier is trained using the kernelized dataset; for instance, we use $k$NN and a Na\"ive Bayes classifiers. Once the classifier is created, labels of unseen samples can be predicted in a similar way, that is, the references are used along with the distance function and the kernel function to map the original object to the kernelized space; the internal classifier is then used to predict the label of the sample using its mapped representation. The rest of this section is dedicated to detail these parameters.

\subsection{Selection of the set of references}
\label{sec:clustering}
We consider four different sampling methods. In any case, we can use samples directly as centers, or we can refine these samples using them as input to compute centroids. We define a center point as an item that is part of the dataset; while a centroid is the geometric mean of a group found by the clustering algorithm. The computation of centroids is straightforward using regions of a Voronoi partition induced by centers. Hereafter, we will use the term references to indicate the use of both centers and centroids. 

\paragraph{Random selection} It is a stochastic algorithm based on taking a random sample, evaluate each configuration, and select the best among the sample. More detailed, we select $R \subset X$ randomly; as commented, it is possible to create a set of centroids computing the nearest neighbor in $R$ of each item in $X$; the geometric mean of those items having $c \in R$ has its nearest neighbor produces the centroid associated to $c$'s region. In some sense, random selection copies the input distribution; however, there is no control about how to handle very dense or very sparse regions.

\paragraph{K-Means} The set of references is computed employing the K-Means clustering algorithm; we use the {\em kmeans++} to select initial centroids (seeds) and reduce intra-cluster variance \cite{arthur2007k}.
We only consider centroid references and the Euclidean distance as the dissimilarity measure for this method.

\paragraph{Density-based selection} This iterative algorithm starts with an empty $R$ and selects a random item in $c \in X$; the set of $\ell$ nearest neighbors of $c$ in $X$ is removed; the procedure repeats while $|X| > 0$. Each $c$ is added into $R$ to create the set of references. Also, the set of nearest neighbors are used to compute the related centroids. The number of references is $k = \lceil X/\ell \rceil$.
This approach is related to Density-net construction; the procedure yields to remove most probable regions first, and continues removing $\ell$ items per iteration; the latter iterations capture and remove least probable regions. Uniformly distributed datasets do not have a preference in the capturing order.

\paragraph{Farthest First Traversal (FFT)} This algorithm is an approximation of the $k$-centers problem and was simultaneously proposed by \cite{gonzalez1985} and \cite{hochbaum1985}; the approximation of FFT is at most two times the optimal solution. The algorithm selects a set of centers $R$ such all items in $R$ are {\em furthest} among them. Since FFT simulates a traversal, each iteration selects an element $c$ as the furthest element from the elements already selected; more detailed, let $r$ be the distant used to select the $i$th element, then a Delone set is formed with the following properties for some distance function $d$:

\begin{itemize}
 \item All centers are separated by at least $r$, i.e., $d(p, q) > r$ for any pair of centers $p, q \in R$.
 \item All objects are covered by some center under $r$ radius, i.e., $d(x, c) \leq r$ for  $x \in X$ and $c \in R$.
\end{itemize}

Let us define $d_{\min}(x)=\min\{d(x,c) \mid c \in R\}$, that is, the distance between $x$ and its nearest center in $R$; then Alg.~\ref{algo:fft} defines the Furthest First Traversal method.

\begin{algorithm}[!ht]
\begin{algorithmic}
\Require {A database $X$ and a distance function $d$}
\Require {The number of centers $k$}
\Ensure {The set of furthest samples $R$}
\State $R \gets \{c\}$ \Comment{A randomly chosen $c$ from $X$}
\While{$|C| < k$}
   \State $w \gets \argmax~\{d_{\min}(x) \mid x \in X \setminus R \}$
   \State $R \gets R \cup \{w\}$
\EndWhile
\end{algorithmic}
\caption{The farthest first traversal algorithm.\label{algo:fft}}
\end{algorithm}

Summarizing, FFT chooses $k$ centers such that the maximum distance from some $x \in X$ to its nearest center in $R$ is minimized; at the same time, it maximizes the inter-cluster distance. Centroid references are generated in the same way that in Density Clustering. Please notice that FFT can use any distance function.

\subsection{An illustrative example for different sampling strategies}
\begin{figure}[!th]
\centering
\subfigure[Ground truth]{
\label{fig:gts}
    \centering\includegraphics[width=0.32\textwidth, trim=1.5cm 1.5cm 1.2cm 1.5cm,clip]{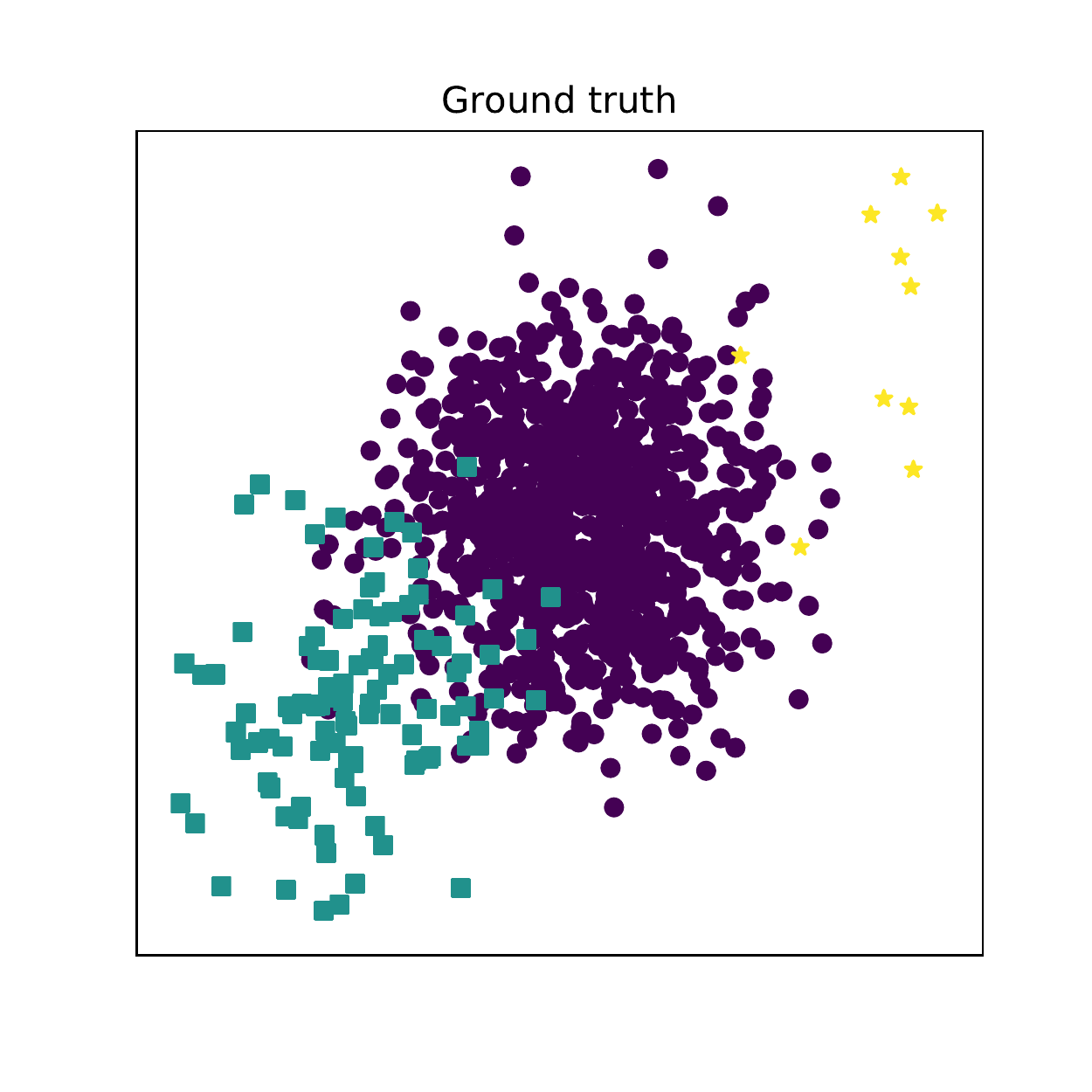}
}\subfigure[K-Means]{
\label{fig:kmeanss}
    \centering\includegraphics[width=0.32\textwidth,trim=1.5cm 1.5cm 1.2cm 1.5cm,clip]{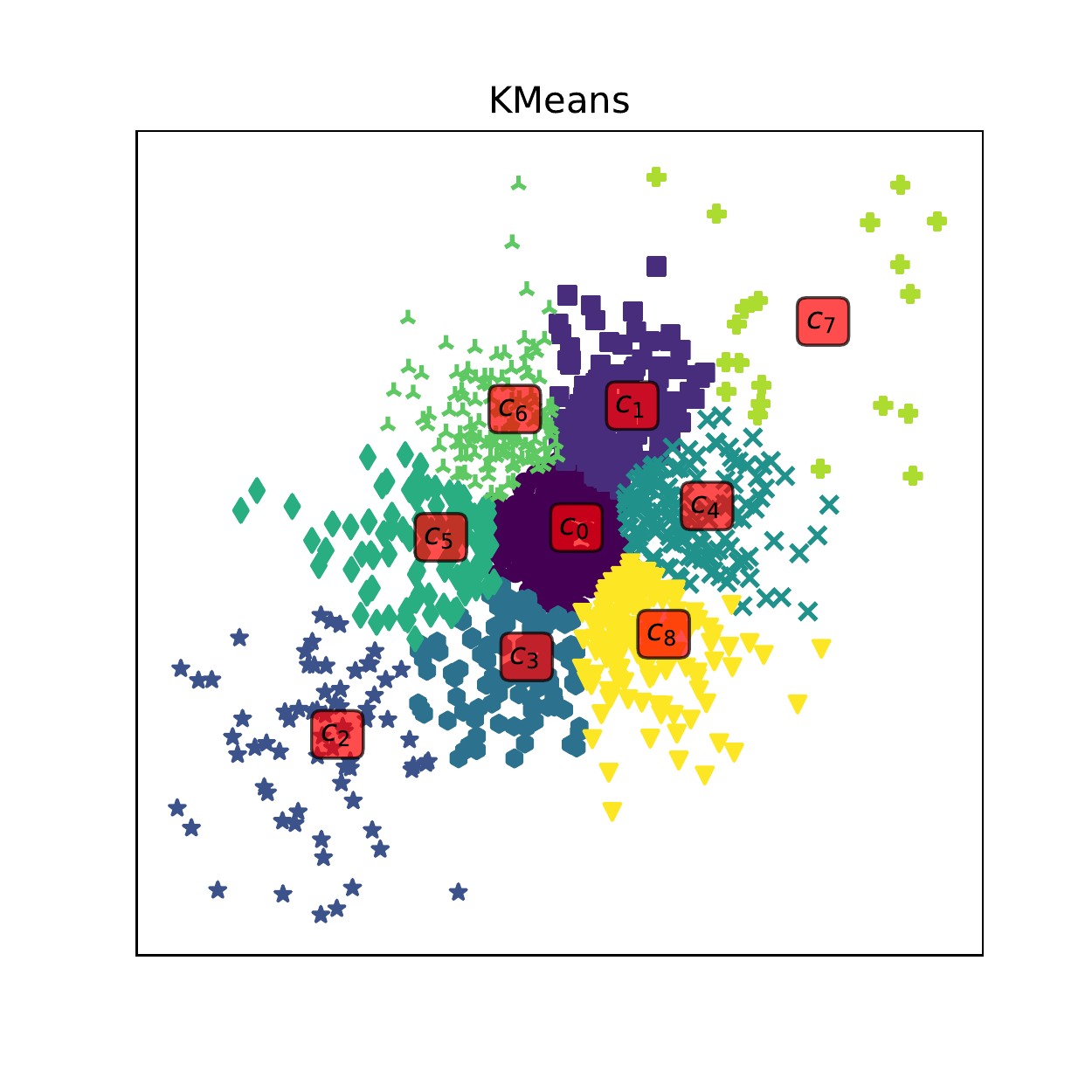}
}\subfigure[FFT]{
\label{fig:ffts}
    \centering\includegraphics[width=0.32\textwidth, trim=1.5cm 1.5cm 1.2cm 1.5cm,clip]{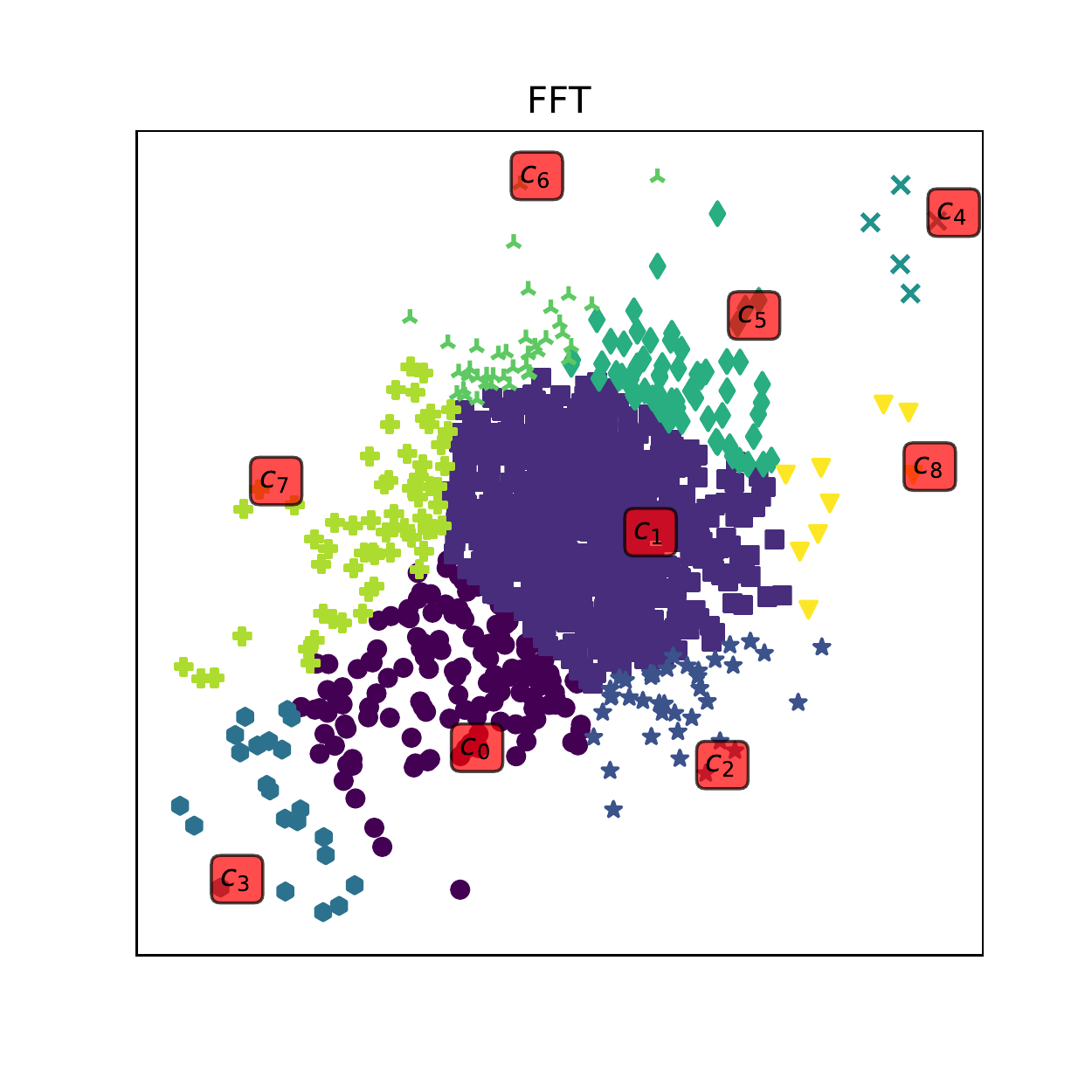}
}

\subfigure[Density]{
\label{fig:dnets}
    \centering\includegraphics[width=0.32\textwidth,trim=1.5cm 1.5cm 1.2cm 1.5cm,clip]{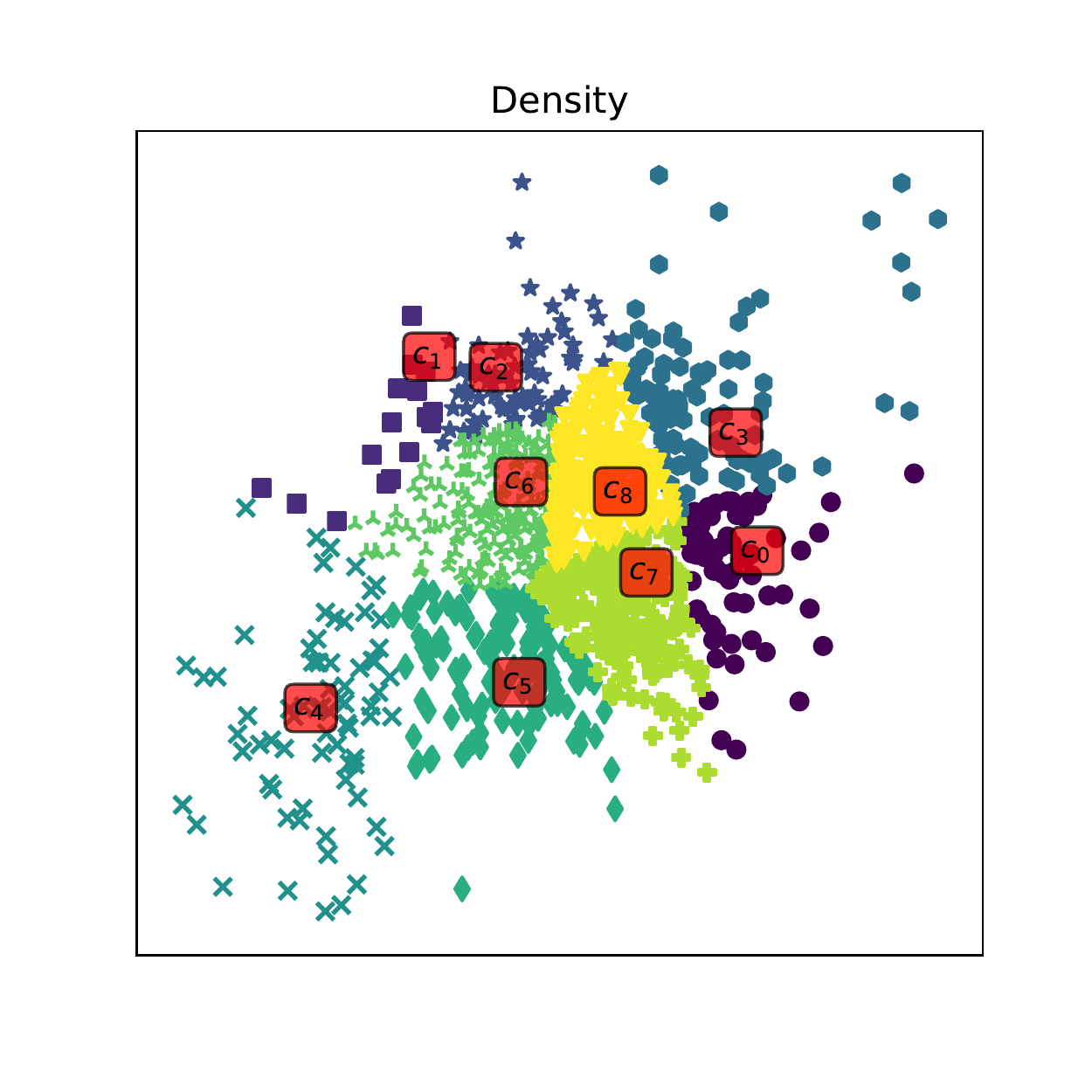}
} \subfigure[Random]{
\label{fig:randoms}
    \centering\includegraphics[width=0.32\textwidth,trim=1.5cm 1.5cm 1.2cm 1.5cm,clip]{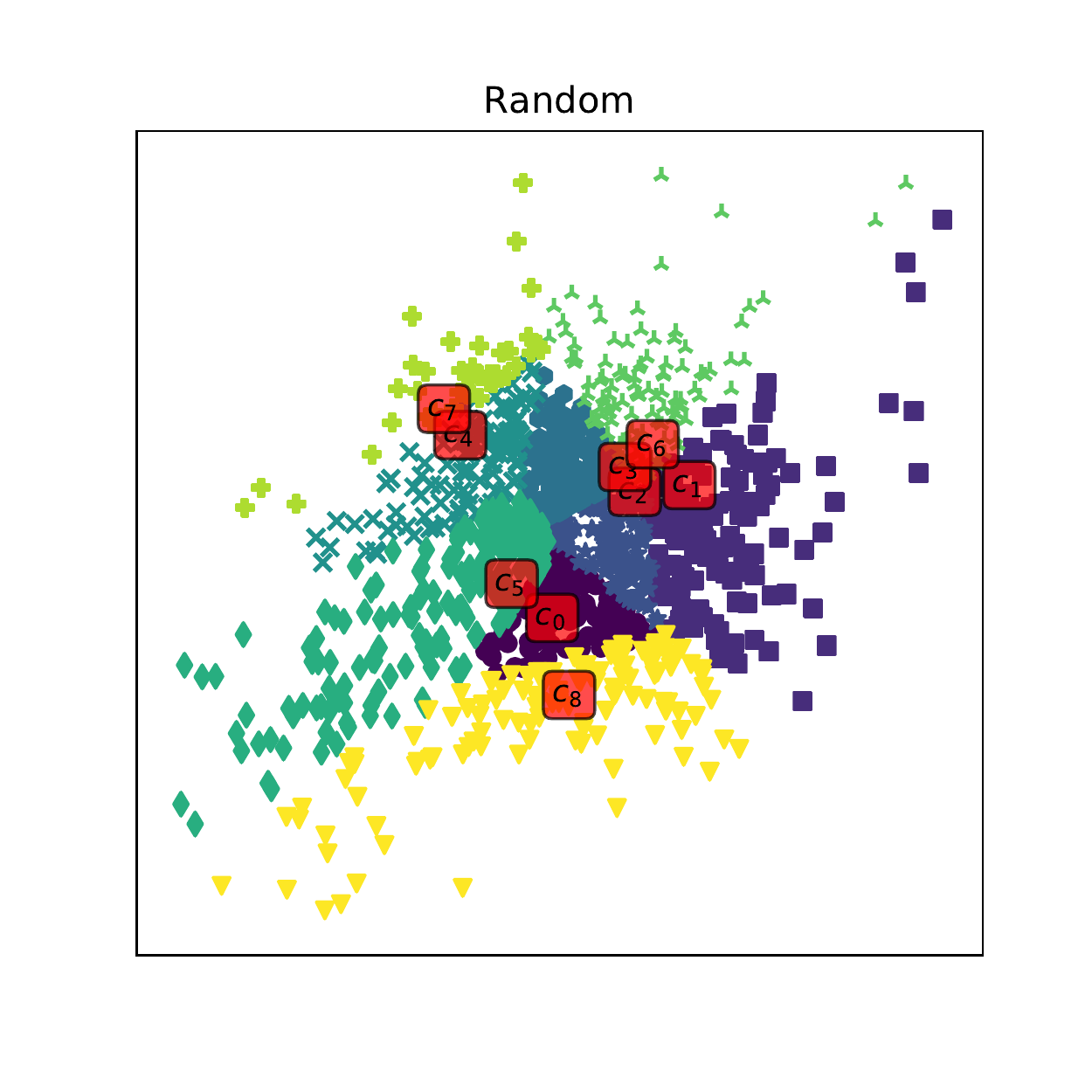}
}
\caption{Prototypes selected from a three cluster toy example by using different sampling strategies; references are drawn as a labeled-red square. Each group is indicated with both a unique marker and a unique color; these regions are computed as the nearest neighbors to centers.}
\label{fig:clusterings}
\end{figure} 

Figures \ref{fig:gts} show a 2-dimensional toy-dataset to develop insight about each sampling strategy; the point collection was generated to have three Gaussian-distributed groups. The figure uses a different color and point marker for each class; also, note that each group has a different number of samples; this situation is similar to some highly-imbalanced datasets. For each sampling method explained in \S\ref{sec:clustering}, we compute nine prototypes based on the Euclidean distance, i.e., ${c_0,\cdots,c_8}$. Note that we are using three times more prototypes than clusters and that groups overlap, also note how each sampling strategy captures different dataset's properties and therefore, they can be used to fit for different tasks.

Figure \ref{fig:kmeanss} illustrate regions generated with K-Means; note that prototypes induced by this strategy are concentrated around the zone with higher mass and evenly distributed around them. This sampling bias can be a problem for unbalanced collections since more dense datasets will be oversampled and low-density regions can remain untouched. Figure \ref{fig:ffts} illustrates the partition induced FFT; please notice how prototypes are evenly distant from each other and around the dataset, independently of their mass-density, maximizing the volume of each region.

Density-based selection, Fig.~\ref{fig:dnets}, is guided by dense zones; as explained in \S\ref{sec:clustering}, it ensures the covering of the entire dataset, but low-density zones will have few prototypes, independently of the volume of each region. Lastly, Fig.~\ref{fig:randoms} shows how random sampling is prone to select prototypes from high-density clouds; note that favoring dense zones may be a desirable feature when it is tried to remove outliers.

Determining the correct strategy is highly dependent on the classification task, i.e., its particular distribution; this is the reasoning behind our model selection scheme.

\subsection{Feature generation}
Once the set of references $R$ is computed, a kernel function $f$ is used to generate the new kernelized feature space $\hat{X}$; that is, we compute $f(x, R)$ over each $x \in X$. More precisely,  $\hat{X}_i = f_k(x_i, R) = \{f(x_i, c_1), \ldots, f(x_i, c_k)\}$, where $f(x_i, c_j)$ is the kernel function. 
The following functions define the kernel functions used in this paper, that is, those which are used for our experiments:

\begin{align}
\textsf{linear}(x, c) & = d(x,c)\\
\textsf{gaussian}(x, c) & =  \exp(-d(x,c) / \sigma_c)\\
\textsf{sigmoid}(x, c) & = \frac{1}{1+\exp(\sigma_c-d(x,c))}\\
\textsf{cauchy}(x, c) & = \frac{1}{1+d(x,c) / \sigma_c}
\end{align}

A reader interested in the properties of these kernel functions is referred to the literature in the field~\cite{genton2001classes,hofmann2008kernel,zhuang2011family,muandet2017kernel}.
Note that these functions are compositions over a distance function $d$ using for two dataset elements $x$ and $c$. Here, $\sigma_c$ is the maximum intra-cluster distance, that is, the maximum distance from center $c$ to any object having $c$ as its nearest neighbor among the set of references; therefore, each region has its own $\sigma_c$ value. Notice that $\sigma_c$ can be set as the last $r$ value for FFT since these references will be evenly distributed.
The feature mapping process is described in Alg.~\ref{alg:feature}; despite its simplicity, it details the requirements and the procedure needed for mapping both training and test datasets.

\begin{algorithm}[ht]
\begin{algorithmic}[1]
\Require {A metric database $X$ and an associated distance function $d$}
\Require {A number of features $k$}
\Require {A set of references $R$ and its associated $\sigma_c$ values}
\Require {A kernel function $f_k$ }
\Ensure {The database $\hat{X}$ defined with the new features space}
 \State $\hat{X} \gets \{~\!\}$
 \For{$x \in X$}
   \State $\hat{x} \gets f_k(x, R)$ \Comment{Both $d$ and $\sigma_c$ are used internally}
   \State $\hat{X} \gets \hat{X} \,\bigcup\, \{\hat{x}\}$
 \EndFor 
\end{algorithmic}
\caption{Generation of new feature space\label{alg:feature}}
\end{algorithm}

\subsection{Internal classifiers}
\label{sec:model-selection}

Our approach uses either $k$NN or Na\"ive Bayes as internal classifiers.
We select $k$NN since it is a well-known non-linear classifier; it is straightforward to use over the new space, and it is a well-known kernel method. It is worth to mention that $k$NN uses a distance function to work; we use Euclidean and Angle distances, but any distance function can be used. The selection of the current distance is independent of that used in previous stages. It is possible to use from one to several neighbors to make the decision; also, we can select to use weight each neighbor in a uniform or differently, for example, based on the distance to the sample being processed.

\subsection{Hyper-parameter optimization}

\begin{table}[ht]
\caption{\label{tb:features} Configuration space of our classification pipeline.}
\centering
\resizebox{.75\textwidth}{!}{
\begin{tabular}{ll}
\toprule
{\bf Name} & {\bf Value} \\ \midrule 
Number of references $k$ & $\{4,8,16,32,64\}$\\
Distance function   & \{Angle, Euclidean\} \\
Sampling method     & \{Density, FFT, K-Means, Random\} \\ 
Kernel function          & \{Linear, Gaussian, Sigmoid, Cauchy\} \\ 
Reference's type        & \{Centers, Centroids\} \\
Internal classifiers    & \{Na\"ive Bayes, $k$NN\} \\
$k$NN weighting scheme  & \{Distance, Uniform\} \\
$k$NN distance function & \{Angle, Euclidean\} \\ 
Number of neighbors     & \{1,5,11,21\} \\
\bottomrule
\end{tabular}
}
\end{table}

Once our pipeline is defined; it is necessary to select the precise algorithm for each classification task. 
As explained in previous sections, we use hyper-parameter optimization to select a competitive classification over a large configuration space; recall that we describe our set of classifiers through its configuration \S\ref{sec:okcc}. Table~\ref{tb:features} summarizes a grid of parameters that define the configuration space. This grid contains more than ten thousand different configurations; however, some of the possible ones are not valid like those having Na\"ive Bayes as its internal classifier and varying $k$NN's parameters. In sum, the configuration space contains close to 4,500 valid configurations. Notice that parameters are represented as a set of categorical values since most parameters are categorical indeed.

The model's performance prediction is computed with a k-folds cross-validation procedure. For instance, our experimental section show results for selecting models using 3-folds. This model validation reduces the chances of a model's overfitting while ensures the selection of a competitive model due to its precise predictions. The selection is led by the balanced error rate (BER) measure. We dubbed this process, and our classifier itself, as Kernel-based Model Selection (KMS).

While the evaluation of the configuration space can be performed in several ways, we decide to use the low-cost Random Search (RS) meta-heuristic over parameter's boundaries defined in Table~\ref{tb:features}. The random search consists of uniformly sampling the configuration space, evaluate the performance of each configuration in the sample, and then select the best performing setup. For instance, our experimental results of the next sections were computed with a sample of 128 configurations.
We decide to perform Grid Search (GS) over the configuration space, i.e., an exhaustive evaluation of the complete grid, with the idea of determining an upper bound of Random search over the defined configuration space.
The interested reader in these meta-heuristics is referred to~\cite{bergstra2012random}.

The exploration of the entire space is prohibitive, due to its size and evaluation's cost for each instance. For instance, each configuration pipeline is evaluated in 4.11 seconds,\footnote{Using our open-source implementation
\url{https://github.com/kyriox/KernelMethods.jl} under CentOS Linux on Intel Xeon E5-2640 v3 @ 2.60GHz.} the time becomes 8.76 minutes for 128 instances for RS and close to six hours for GS. Even when it is possible to reduce this cost with parallel and distributed computation, it is worth to mention that RS achieves competitive results and even when it is possible to improve its performance it becomes a simple and effective alternative to other costly methods like GS, and also being susceptible to be optimized for high-performance computing environments. Other meta-heuristics may be of use on more overwhelming combinations.


\subsubsection{Improving performance via ensembling}
\label{sec/ensemble}
To stabilize and improve the performance, we ensemble a group of KMS instances into a KMS ensemble (KMSE); the ensemble is also used to avoid overfitting since several configurations are harder to overfit.
The procedure requires the selection of a group of models having a proved high performance; therefore, we select top-performing classifiers under the random search application.
While this procedure keeps the ensembling method simple, we also maintain a construction cost low since the construction time remains almost identical to any of our hyper-heuristic optimization schemes.

In particular, our ensemble implementation uses a voting scheme to determine the label of new samples. A procedure to determine the size of the ensemble is studied in \S\ref{sec/ensemble-size}.

%
%

\section{Experimental results}
\label{sec:results}
This section is dedicated to characterize and experimentally prove the performance of our classifiers. Our testing methodology is composed of two stages.
The first one characterizes our method using nine datasets for this purpose. The nine databases used to tune our methods, more precisely, to select the size of our ensemble and some of the numerical limits presented in Table~\ref{tb:features}. These benchmarks were obtained from the Gunnar Raetsch's collection;\footnote{Available at \url{http://theoval.cmp.uea.ac.uk/matlab/default.html#benchmarks}} Table~\ref{tb:datasets} shows their characteristics. These benchmarks are binary problems (2-classes) and have a relatively low number of training samples. Despite these limitations, these datasets have been widely used to measure the performance of classification methods in the literature. Please note that each of these benchmarks has 100 training and test splits, except for the {\em image} benchmark which has only 20 splits. Following the literature standard, we work with average measures over these splits.

The second stage is dedicated to validation; the core idea is to use an independent collection of datasets to validate the decisions that were made in the first stage. Table~\ref{tb:datasets} also describes these benchmarks; we can observe that the number of classes varies from 2 to 26 and that the number of dimensions spans from 4 to 170. Moreover, the number of samples is also more varied and more extensive than those found in first-stage benchmarks.
Most of these datasets were collected from UCI's Machine Learning Repository\footnote{\url{https://archive.ics.uci.edu/ml/datasets.php}} with the exception of {\em semeval} and {\em tassgeneralcorpus}. These datasets were generated from the datasets provided for two Twitter Sentiment Analysis challenges, namely, TASS'16 (Spanish Sentiment Analysis, General Corpus \cite{tass2017}) and SemEval'2017 (Task 4: English Sentiment Analysis \cite{semeval2017}). Text's feature vectors were computed using the fastText tool.\footnote{\url{https://github.com/facebookresearch/fastText}}


\begin{table}[!h]
\centering
\caption{\label{tb:datasets} Brief description of the databases used during the characterization and validation stages.}
\resizebox{0.7\textwidth}{!}{
\begin{tabular}{lrrrrr}
\toprule
Name & Classes & Dimension & Samples & Training & Test  \\ \midrule
\multicolumn{5}{c}{Characterization datasets} \\ \midrule
banana    & 2       & 2  & 5,300 & 400  & 4,900 \\
thyroid   & 2       & 5  & 215  & 140  & 75 \\
diabetes  & 2       & 8  & 768  & 468  & 300 \\
heart     & 2       & 13 & 270  & 170  & 100 \\
ringnorm  & 2       & 20 & 7,400 & 400  & 7,000 \\
twonorm   & 2       & 20 & 7,400 & 400  & 7,000 \\
german    & 2       & 20 & 1,000 & 700  & 300 \\
image     & 2       & 20 & 2,310 & 1,300 & 1,010 \\
waveform  & 2       & 21 & 5,000 & 400  & 4,600 \\ 
\midrule
\multicolumn{5}{c}{Validation datasets} \\
\midrule
agaricuslepiota   & 7  & 22   &   8,124 &   5,686 &   2,438\\ 
apsfailure        & 2  & 170  &  76,000 &  60,000 &  16,000  \\
banknote          & 2  & 4    & 1,372 &    960 &    412 \\
bank              & 2  & 16   & 45,211 &  31,647 &  13,564 \\
biodeg            & 2  & 41   &  1,055 &    738 &    317\\
car               & 4  & 6    &   1,728 &   1,209 &    519\\ 
censusincome      & 2  & 41   & 299,285 & 199,523 &  99,762  \\
cmc               & 3  & 9    &  1,473 &   1,031 &    442\\
drugconsumption   & 7  & 30   & 1,885 &   1,319 &    566\\
indianliverpatient& 2  & 10   &    582  &    407 &    175 \\ 
iris              & 3  & 4    &     149 &    104 &     45 \\
krkopt            & 18 & 6    &  28,056 &  19,639 &   8,417 \\
letterrecognition & 26 &  16   & 20,000 &  14,000 &   6,000  \\
mlprove           & 2  & 56   &  6,118  &   4,588 &   1,530 \\
musk1             & 2  & 166  &    476  &    333  &    143 \\
musk2             & 2  & 166  &   6,598 &  4,618  &   1,980 \\
optdigits         & 10 & 64   &   5,620 &   3,823 &   1,797 \\ 
pageblocks        & 5  & 10   &   5,473 &   3,831 &   1,642 \\
parkinson         & 2  & 22   &  194    &    135 &     59 \\
pendigits         & 10 & 16   & 10,992  &   7,494 &   3,498\\
segmentation      & 7  & 19   &   2,310 &    210 &   2,100 \\
semeval           & 3  & 100  & 58,032  & 45,748 & 12,284 \\
tae               & 3  & 5    &   150 &    105 &     45 \\
tass              & 4  & 100  & 68,016  & 7,218 & 60,798 \\
yeast             & 10 & 9    & 1,484 &   1,038 &    446\\ 
\bottomrule
\end{tabular}
}
\end{table}

\subsection{Determining the size of the ensemble}
\label{sec/ensemble-size}
\begin{figure*}[th!]
\centering
\includegraphics[width=0.96\textwidth, trim={3cm 0cm 3cm 1cm}, clip]{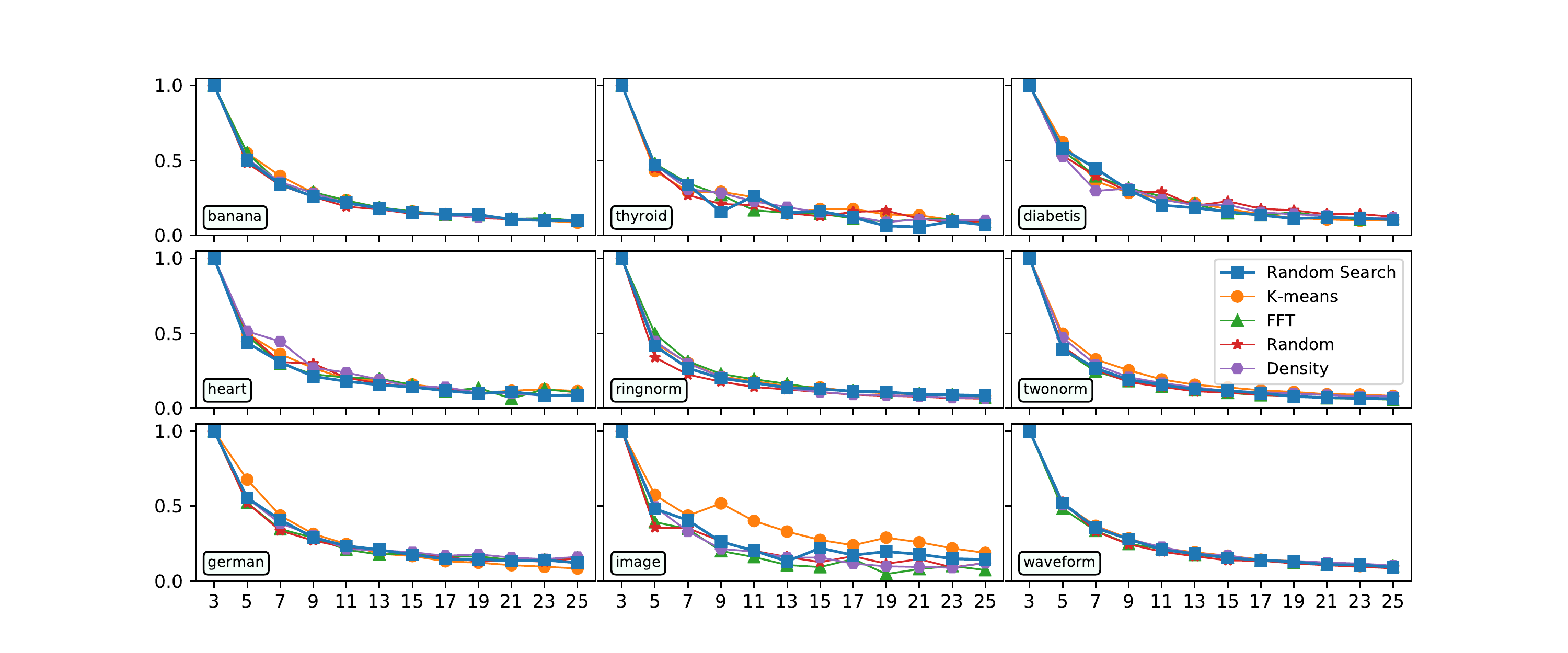}
\caption{\label{fig:ensk} Consensus ratio between ensembles of size $\ell$ and $\ell+2$. Please recall that we always use top instances to build the ensemble.} 
\end{figure*}

As described in \S\ref{sec/ensemble}, we use ensembles to improve, stabilize the expected performance, and avoid overfitting. Our KMSE includes the selection of top-$\ell$ performing instances, measured in training set. In particular, we use the already evaluated classifiers in the Random search procedure; therefore, the cost of creating a KMSE is almost identical to that achieved with the Random search method.  The overall prediction is made with a majority voting scheme; in case of ties, the prediction is randomly selected among most voted predictions.

Instead of selecting the ensemble's size based on an additional cross-validation stage, we determine it using a consensus scheme. More detailed, the consensus is measured as the agreement in the label's predictions between a KMSE with top-$\ell$ instances, and those labels predicted with a KMSE with top-$(\ell+i)$ instances. 
Figure~\ref{fig:ensk} illustrates the consensus of KMSE on our nine characterization benchmarks; in particular, the figure starts on $\ell=3$ and fixes $i=2$. Each curve is the proportion of discordant predictions KMSE with different sampling methods, namely, K-means, FFT, Random, and Density-based selection. On the other hand, KMSE with Random search allows the selection of any sampling methods. All discordant ratios were normalized by the maximum ratio per benchmark to obtain values between 0 and 1.

Figure~\ref{fig:ensk} shows how small values of $\ell$ produce the most significant differences among predictions. It is worth to mention that prediction's computational cost is tightly linked to $\ell$ and we shall select a value that ensures low variance in the predicted labels but also low cost in the prediction procedure. Based on the figure, $\ell$ should be between 9 and 25; these values yield to a consensus with relatively low computing cost. We decided to use $\ell=15$ since it has stable consensus in almost all characterization benchmarks; the performance is compared in the following paragraphs.

\subsection{The performance of KMS classifiers}
\label{sec/ensemble-performance}

\begin{table}[!ht]
\caption{\label{tb:summary} BER performance comparison of different sampling strategies for the {\bf KMS} and KMSE on our characterization benchmarks; ensembles use $\ell=15$ instances. The last row summarizes the performance of the method with the average rank over benchmarks, note that methods (columns) are sorted by average rank. Best values per row are marked in bold.}
\resizebox{\linewidth}{!}{
\begin{tabular}{l rrrr rr|rr rrrr}
\toprule

{Benchmark} & KMSE  & KMSE  & KMSE  & KMSE  & KMSE  & KMSE & KMS  & KMS  & KMS  & KMS  & KMS & KMS  \\
                 & GS & RS & FFT & Random & Density & K-Means & RS & FFT & Density & GS & K-Means & Random \\ \midrule
{banana} & 11.87 & 11.84 & 11.88 &\bf 11.84 & 11.91 & 11.87 & 12.11 & 12.26 & 12.45 & 12.25 & 12.18 & 12.37 \\
{thyroid} &\bf 5.25 & 5.68 & 5.71 & 5.51 & 5.75 & 5.55 & 6.35 & 5.87 & 6.57 & 6.12 & 5.93 & 6.07 \\
{diabetis} & 28.57 & 28.81 &\bf 28.35 & 28.99 & 28.46 & 30.35 & 29.17 & 28.63 & 28.76 & 29.34 & 31.62 & 29.83 \\
{heart} & 17.50 & 17.42 &\bf 17.31 & 17.75 & 17.64 & 17.90 & 19.28 & 18.00 & 18.44 & 18.69 & 18.56 & 18.28 \\
{ringnorm} & 1.60 & 1.62 & 2.42 & 1.98 &\bf 1.56 & 2.11 & 1.98 & 2.85 & 1.87 & 1.87 & 2.30 & 2.36 \\
{twonorm} & 2.39 & 2.38 & 2.39 &\bf 2.38 & 2.39 & 2.42 & 2.77 & 2.81 & 2.68 & 2.83 & 2.63 & 2.76 \\
{german} & 27.70 & 27.83 &\bf 27.69 & 27.87 & 27.88 & 36.10 & 28.59 & 28.60 & 28.63 & 28.89 & 35.33 & 28.66 \\
{image} & 4.11 & 4.14 &\bf 4.02 & 4.25 & 4.36 & 5.45 & 4.25 & 4.18 & 4.63 & 4.30 & 5.25 & 4.47 \\
{waveform} & 9.69 & 9.68 &\bf 9.66 & 9.68 & 9.69 & 9.75 & 10.40 & 10.57 & 10.26 & 10.47 & 10.42 & 10.52 \\
{\bf avg. rank} &\bf 2.89 & 3.11 & 3.33 & 3.78 & 4.44 & 7.44 & 8.22 & 8.22 & 8.33 & 9.22 & 9.44 & 9.56 \\
\bottomrule
\end{tabular}}
\end{table}

Table~\ref{tb:summary} shows the average BER for our characterization benchmarks.
Models were selected in the training set using the specified method, and the error is measured in the test set. The table shows two kinds of values, average BER values, and average rank positions; methods are ordered by the average rank achieved by each method in all benchmarks. Please recall that lower BER values are desired; also, we desire lower average ranks since the best possible rank is 1. 

The table lists the performance of both single-instance methods and ensemble methods. Ensemble methods listed in the table have fifteen instances under a voting scheme; these instances were selected as the best-performing ones in training set among the evaluated configurations. In particular, both RS and GS perform the optimization taking into account all sampling methods while methods listed as a single sampling method perform an RS fixing the sampling method.

Table \ref{tb:summary} shows that methods based on ensembling outperform, consistently, single instance classifiers; the precise method to determine the size of the ensemble is detailed in \S\ref{sec/ensemble-size}. The best ensemble method is found with GS, followed by RS. On the third place, we found FFT which achieves five best positions for {\em diabetis}, {\em heart}, {\em german}, {\em image}, and {\em waveform}. FFT could be the best method, but it has a poor performance on {\em ringnorm}. So, here resides the power of GS and RS since both explore the entire set of parameters and can turn around when bad cases arise for a good method. On the rest of the benchmarks, Random selection achieves two best places for {\em banana} and {\em ringnorm}; Density selection performs the better for {\em ringnorm} and GS for {\em thyroid}. Please recall that all best places arise on ensemble methods. It is worth no mention that GS obtained just one best place and RS none of them, and both occupy the best positions on the global ranking due to their competitive and low variant performance.

\begin{table}[!t]
    \centering
\caption{
BER performance comparison of different sampling strategies for the {\bf KMS} and KMSE on our validation benchmarks; ensembles use $\ell=15$ instances. The last row summarizes the performance of the method with the average rank over benchmarks, note that methods (columns) are sorted by average rank. Best values per row are marked in bold.
    }
\label{tb:summaryv}
\resizebox{\linewidth}{!}{
\begin{tabular}{lrrrrrrrrrr}
\toprule
{Benchmark} &            KMS &   KMSE &  KMS     &   KMSE & KMS & KMS    &   KMSE  &  KMSE  & KMS & KMSE \\
{} &            RS   &   RS  &  Density &   FFT  & FFT & Random & Density & Random & K-Means &    K-Means \\
\midrule
{agaricuslepiota}   &          40.60 &          37.66 &  \textbf{34.16} &          37.62 &          36.54 &          40.64 &          38.77 &          38.80 &          38.27 &          50.55 \\
{apsfailure}        &           7.21 &           6.97 &   \textbf{6.54} &           7.30 &           7.20 &           6.66 &           6.72 &           7.18 &           6.57 &           8.48 \\
{bank}               &          29.85 &  \textbf{29.33} &          30.25 &          38.69 &          41.34 &          37.59 &          29.57 &          38.99 &          39.22 &          46.05 \\
{banknote}           &           0.21 &           0.21 &   \textbf{0.00} &           0.29 &           0.21 &           0.58 &           0.50 &           1.08 &           0.21 &           2.71 \\
{biodeg}             &          25.26 &          24.71 &          24.58 &          23.83 &          23.83 &  \textbf{23.09} &          23.81 &          24.90 &          28.68 &          28.87 \\
{car}                &  \textbf{33.63} &          35.68 &          39.00 &          37.06 &          41.66 &          39.74 &          35.60 &          37.73 &          41.51 &          45.62 \\
{censusincome}      &          31.79 &          32.40 &  \textbf{30.30} &          32.27 &          32.15 &          32.52 &          32.43 &          33.10 &          37.23 &          39.84 \\
{cmc}                &          50.15 &          47.56 &          48.43 &          47.03 &          46.94 &          48.08 &          48.45 &  \textbf{46.84} &          51.90 &          53.77 \\
{drugconsumption}   &          71.55 &          72.81 &          72.71 &          72.02 &          72.04 &          72.67 &          73.57 &          73.22 &          71.80 &  \textbf{70.13} \\
{indianliverpatient} &          34.47 &          34.79 &          34.15 &          34.91 &          35.52 &          36.43 &  \textbf{33.74} &          35.52 &          34.28 &          34.91 \\
{iris}               &           2.08 &           2.08 &           2.08 &           2.08 &           2.08 &   \textbf{0.00} &           2.08 &           2.08 &           5.79 &           3.94 \\
{krkopt}             &          58.84 &          62.90 &          62.28 &          61.38 &  \textbf{53.49} &          61.81 &          66.09 &          68.95 &          67.27 &          71.47 \\
{letterrecognition} &           9.11 &           9.41 &          10.09 &   \textbf{8.26} &           8.40 &          10.38 &          11.28 &          11.27 &          15.53 &          36.93 \\
{mlprove}           &           2.90 &           4.45 &   \textbf{2.01} &           5.55 &           5.09 &           2.58 &           3.93 &           4.18 &           8.20 &          18.48 \\
{musk1}              &          16.96 &  \textbf{12.88} &          13.49 &          16.14 &          16.73 &          21.25 &          17.80 &          15.53 &          17.57 &          17.99 \\
{musk2}              &           6.41 &           8.18 &           5.99 &           9.23 &           6.35 &           5.00 &           8.95 &           8.85 &   \textbf{4.30} &          13.38 \\
{optdigits}          &   \textbf{3.69} &           3.69 &           3.69 &           3.87 &           3.76 &           4.03 &           4.20 &           5.26 &           4.15 &           7.14 \\
{pageblocks}        &          14.94 &          17.58 &          25.51 &          18.27 &  \textbf{14.45} &          23.05 &          22.71 &          22.38 &          51.04 &          53.25 \\
{parkinsons}         &  \textbf{27.33} &          30.45 &          29.29 &          33.58 &          30.45 &          35.90 &          30.45 &          33.58 &          30.45 &          32.41 \\
{pendigits}          &           3.14 &           3.28 &           4.45 &   \textbf{2.94} &           3.32 &           3.83 &           4.14 &           4.15 &           3.58 &           6.05 \\
{segmentation}       &          21.86 &          20.14 &          21.24 &          18.43 &          18.90 &  \textbf{18.10} &          19.90 &          22.19 &          20.57 &          24.57 \\
{semeval}       &          37.59 &          36.98 &          37.05 &          37.16 &          37.20 &  \textbf{36.97} &          37.09 &          37.05 &          37.64 &          37.68 \\
{tae}                &          42.93 &          43.43 &          48.99 &          41.92 &          52.02 &          49.49 &          48.99 &          47.47 &  \textbf{38.89} &          59.09 \\
{tass}          &          46.32 &          46.58 &          47.10 &  \textbf{46.06} &          46.10 &          48.18 &          47.51 &          47.62 &          49.66 &          49.68 \\
{yeast}              &          69.53 &          68.09 &          67.19 &          66.40 &          69.05 &          70.06 &          69.97 &  \textbf{66.28} &          70.11 &          69.93 \\
\textbf{avg. rank}       &                \textbf{4.04} &                \textbf{4.04} &      4.08 &           4.40 &           4.56 &           5.68 &           5.84 &           6.48 &           6.68 &           9.20 \\
\bottomrule
\end{tabular}}
\end{table}

Please note that the selection of the ensemble's size $t$ was made with our characterization dataset. On the other hand, Table~\ref{tb:summaryv} lists the performance of the same methods (excepting for GS based methods, due to its high computational cost) on the validation benchmarks, ensembles fix $t=15$. Unlike our previous experiment, ensemble methods do not surpass all single instance methods. This performance indicates that $t$ must be adapted for the precise benchmark and that the homogeneity of the characterization datasets where the cause of the performance's domination. However, it is possible to find a pattern on the impact of sampling methods, both RS and FFT excel on its performance while our methods barely take advantage of samples produced by K-Means. Note that KMS Density and KMS Random, the first performs best on five benchmarks and the second performs best on four benchmarks; their position in the global ranking is surpassed by methods that achieve three and two best places (methods with best two positions in average rank). This behavior suggests that KMS Density and KMS Random produce a higher variance in its performance.

It is worth to notice that both ensembles and single-instance classifiers perform pretty similar if they have the same sampling method. Since ensemble-based classifiers do not dominate as in Table~\ref{tb:summary}, therefore, the ensembling process must be refined for adapting better to the dataset being processed.

\subsection{Comparison among different alternatives}
\label{sec:comparison-with-alternatives}

\begin{figure*}[!ht]
\centering
\includegraphics[width=\textwidth]{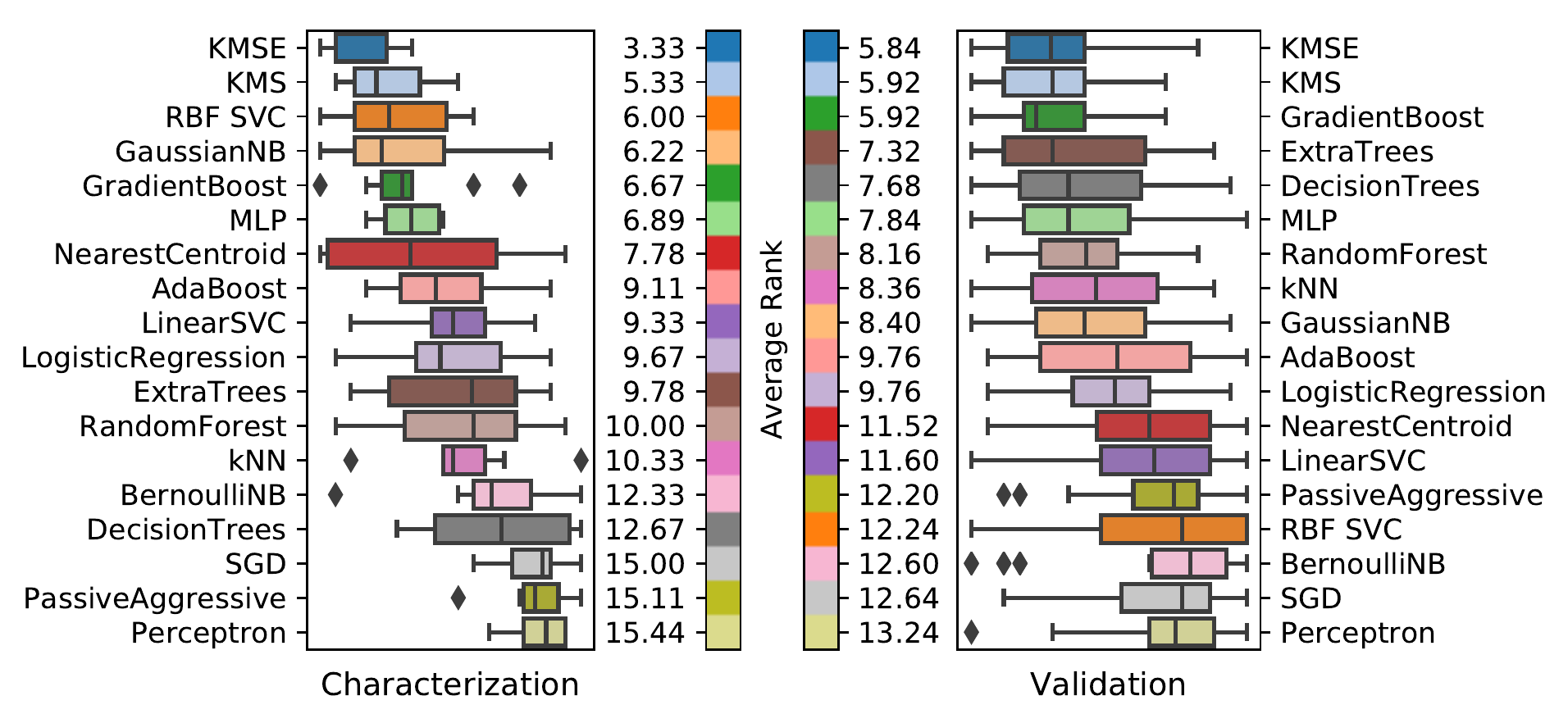}
\caption{Performance comparison based on the average rank over the nine characterization benchmarks (left) and the 25 validation benchmarks (right). \label{fig:avgRank}}
\end{figure*}

Figure~\ref{fig:avgRank} compares the BER performance of our contributions with sixteen popular classifiers implemented in the widely used scikit-learn~\cite{sklearn}.
For this experiment, we consider only KMS RS and KMSE RS, named from now on as  KMS and KMSE, for the sake of simplicity. Methods are presented in ascending average rank, so smaller is better; the box-plots show the distribution of ranks per benchmark, per method. Box's colors are fixed in both figures to simplify the tracking of methods on both sides; please note that each side has a different scale of ranks.
The left side of the figure shows the performance on our characterization datasets where we tune our configuration space, i.e., select the size of ensembles and the limits of its numeric hyper-parameters. We can observe that KMSE achieves the best performance, and KMS achieves the second place. It is worth to notice that variance is relatively small for these methods. The Support Vector Machines with RBF kernel achieves the third better average rank, followed y Gaussian Na\"ive Bayes and Gradient Boosting. Note that the later has a small variance. Note that $k$NN performs relatively bad since it achieves the 13th position on this rank; please recall that KMS and KMSE use $k$NN and Gaussian Na\"ive Bayes after kernel projections; therefore, our procedure improves over them. A Wilcoxon signed-rank test over BER performances shows that KMSE is statistically different to all methods ($p$-values below 0.01) while KMS performs similar to RBF SVC and Gaussian Na\"ive Bayes having $p$-values of 0.23 and 0.15, respectively.

On the right side of Figure~\ref{fig:avgRank}, the performance of KMSE and KMS achieve best ranks for validation benchmarks; here, Gradient Boosting achieves the third better place, followed by Extra trees and Decision Trees, MLP, and Random Forest; except for MLP, most of them are different kinds of decision trees. Both $k$NN and Gaussian Na\"ive Bayes achieve eighth and ninth positions, again, our KMS and KMSE methods improve significantly over using raw input, i.e., without kernel transformations. In contrast, Support vector machines with RBF kernel passes from the third position on the characterization datasets to the fifteenth position in validation benchmarks. 
A Wilcoxon signed-rank test over BER performances show that KMS and KMSE are statistically different, i.e., a $p$-value of 0.001; on the other side, KMS is similar to Gradient Boosting with a $p$-value of 0.16.

\subsection{The effect of input's normalization}
\begin{table*}[!ht]
    \centering
    \caption{Average Rank for KMS, KMSE, and popular classifiers when using different scaling strategies. Rows are sorted by the last column, the grand average rank (i.e., the mean of average ranks). The raw column also presents ranks inside the parenthesis.}
    \label{tab:avgRankNorml}
    \resizebox{\linewidth}{!}{
    \begin{tabular}{lrrrrrrr}
\toprule
{} &   Raw &  Standarization &  MinMax &  MaxAbs &  Quantile &  Yeo-Johnson &  {\bf Grand } \\
{} &    &   &   &   &   &   &  {\bf Avg. Rank} \\
\midrule
KMS &   $5.92_{(2)}$ &       5.92 &    5.32 &    {\bf 5.04} &      {\bf 5.16} &         {\bf 4.68} &       {\bf 5.34} \\
MLP               &  $7.84_{(6)}$ &{\bf 5.52} &  {\bf 5.24} &    5.44 &      5.20 &         5.20 &       5.74 \\
KMSE               &  ${\bf 5.84}_{(1)}$ &  5.84 &    6.52 &    5.96 &      5.88 &         5.72 &       5.96 \\
GradientBoost      &  $5.92_{(3)}$ &        6.80 &    6.64 &    7.00 &      6.84 &         7.00 &       6.70 \\
ExtraTrees         &  $7.32_{(4)}$ &        7.88 &    8.24 &    8.40 &      8.76 &         8.24 &       8.14 \\
DecisionTrees      &  $7.68_{(5)}$ &        8.72 &    8.84 &    9.08 &      8.88 &         8.84 &       8.67 \\
RandomForest       &  $8.16_{(7)}$ &        8.72 &    9.20 &    8.60 &      9.08 &         9.16 &       8.82 \\
kNN                &  $8.36_{(8)}$ &        9.08 &    9.16 &    9.40 &      9.24 &         9.48 &       9.12 \\
RBF SVC            & $12.24_{(15)}$ &        8.68 &    8.68 &    8.72 &      8.92 &         8.60 &       9.31 \\
NearestCentroid    & $11.52_{(12)}$ &        9.64 &    9.60 &    9.76 &      9.72 &         9.84 &      10.01 \\
GaussianNB         &  $8.40_{(9)}$ &        10.52 &   10.32 &   10.68 &     10.64 &        10.44 &      10.17 \\
LogisticRegression &  $9.76_{(10)}$ &        10.44 &   10.56 &   10.32 &     10.48 &        10.28 &      10.31 \\
LinearSVC          & $11.60_{(13)}$ &        10.12 &   10.20 &   10.48 &     10.16 &        10.52 &      10.51 \\
AdaBoost           &  $9.76_{(11)}$ &        11.08 &   11.20 &   11.36 &     11.28 &        11.36 &      11.01 \\
Perceptron         & $13.24_{(18)}$ &        12.28 &   12.20 &   12.52 &     12.28 &        12.44 &      12.49 \\
BernoulliNB        & $12.60_{(16)}$ &        12.76 &   12.68 &   12.92 &     12.72 &        12.84 &      12.75 \\
PassiveAggressive  & $12.20_{(14)}$ &        13.96 &   13.44 &   12.56 &     12.96 &        11.88 &      12.83 \\
SGD                & $12.64_{(17)}$ &        13.04 &   12.96 &   12.76 &     12.80 &        14.48 &      13.11 \\
\bottomrule
\end{tabular}}

\end{table*}
 
Since our methods are based on kernel functions, their performance is driven by the effectiveness of the selected distance function. Therefore, the distribution and scale of each variable may determine the performance since distance functions; for instance, a variable with a significantly bigger scale may dominate the final distance values hiding useful information from rest of the variables.

In this experiment, we compare the performance impact of different techniques to scale the input's data. The experiment reports results for our contribution and several popular classifiers; explained in the previous experiment. More precisely, we test the following scalers:
\begin{itemize}
\item {\em Standardization} is perhaps one of the most common transformations applied for any machine learning user; the process removes the mean and scales data to have unit variance.
\item {\em MinMax} that scales each feature between a range defined by a minimum and maximum value.
\item {\em MaxAbs} scales each feature by dividing by its maximum absolute value.
\item {\em Quantile} transforms features by using quantile information.
\item {\em Yeo-Johnson} transformer, which is part of the power transformers family whose aim to improve the normality and symmetry of the data \cite{yeo2000new}.
\end{itemize}

Table \ref{tab:avgRankNorml} shows the average rank for each one of the evaluated classifiers in validation benchmarks, using BER as the ranking score. The table contains the average rank achieved for each method over raw input and the five scaling as mentioned above methods; the last column shows the average rank obtained from the performance achieved on different scaling methods (including the raw input). Note that this experiment compares using average values and presents an average rank with them, i.e., the grand average rank; these results help us in determining the robustness of a method to input data and how the input can be preprocessed to improve the final classification. It is not intended to compare the performance on single benchmarks.

As illustrated in Table~\ref{tab:avgRankNorml}, our KMS achieve the best result, followed by MLP, KMSE, and Gradient Boosting. KMS achieve three best positions for MaxAbs, Quantile, and Yeo-Johnson, while keeping an excellent performance for all scalers. Despite that KMSE achieves the best performance on raw input, it is placed on third-best rank for grand averaging; a wrong ensemble size may cause this performance. Note that MLP takes advantage of scaling, going from the sixth rank in raw to a grand average rank of two; it also achieves two best positions using Standardization and MinMax scalers. Gradient boosting changes barely its global rank, and it goes from a third rank to a fourth because of MLP's improvement. In this experiment, $k$NN and Gaussian Na\"ive Bayes has a relatively bad performance; it is worth to recall that our KMS is based on these classifiers once kernel mapping is performed; this significant performance's difference evidence that our contribution, as a whole, improves its parts.

A Wilcoxon signed-rank test shows that, based on average ranks, KMS, KMSE, and MLP have similar performance; for instance, KMS and MLP have a $p$-value of 0.22, and KMS and KMSE a $p$-value of 0.08. KMSE and MLP also share a high $p$-value of 0.30. It is worth to notice that MLP and Gradient Boosting also have a high $p$-value of 0.30. Beyond mentioned $p$-values for KMS, KMSE, MLP, and Gradient Boosting, other combinations (among them and other methods) yield to $p$-values smaller than 0.05.
Therefore, even when KMS and KMSE excel in their performance, both are statistically similar to other popular classifiers when we use different scalers, at least on our validation benchmarks.

\section{Conclusions}
\label{sec:conclusions}

This work introduces a new family of classifiers based on projecting the input space into a projection based on different sampling strategies and kernel functions. We call these family as Kernel-based Model Selection (KMS). The idea is to find a projection of input data where typical $k$NN and Na\"ive Bayes methods improve their performance significantly.
Therefore, our methods are simple to implement and have an excellent performance, based on the experimental evidence.
Our contribution is an entire pipeline that adapts to a wide range of problems, from linear to non-linear tasks. The pipeline is composed of a distance function, a sampling method, a kernel function, and a simple classifier working on the mapped space. We experimentally show that some sampling strategies adapt better for different problems. 

We ensembled several KMS instances to compose our KMSE; this ensemble-based classifier may improve over single instance classifiers and also performs competitively with other industrial-strength alternatives on our benchmarks. Even when the prediction cost is increased proportionally to the ensemble's size, the construction cost is almost similar to the single-instance KMS. However, it is necessary to mention KMSE's performance is linked to its parameters, i.e., the size of the ensemble and the summarizing algorithm (simple voting for our implementation), both parameters are fixed for our contribution. It is necessary to find better algorithms to fit these hyper-parameters to individual tasks.

We validated our claims experimentally by using a characterization set of benchmarks and a validation set too. The idea of using two kinds of benchmarks comes from the necessity of finding evidence of generalization through removing the bias induced by hyperparameter tuning, so we only touch characterization benchmarks for the design of the hyper-parameter limits and values. The results show that our methods are competitive, in both characterization and validation benchmarks under average BER as compared with a wide range of industrial-strength classification methods like Gradient Boosting, Neural nets, SVM, AdaBoost, Random Forest, among others available in the scikit-learn package. For instance, we found that our methods have the best mean rank among all compared methods under our benchmarks. 

\section*{References}
\small
\bibliographystyle{plain}
\bibliography{bibtex}

\begin{thebibliography}{10}

\bibitem{arthur2007k}
David Arthur and Sergei Vassilvitskii.
\newblock {K-means++: The advantages of careful seeding}.
\newblock In {\em Proceedings of the eighteenth annual ACM-SIAM symposium on
  Discrete algorithms}, pages 1027--1035. Society for Industrial and Applied
  Mathematics, 2007.

\bibitem{baudat2003feature}
Gaston Baudat and Fatiha Anouar.
\newblock {Feature vector selection and projection using kernels}.
\newblock {\em Neurocomputing}, 55(1-2):21--38, 2003.

\bibitem{belkin2003laplacian}
Mikhail Belkin and Partha Niyogi.
\newblock {Laplacian eigenmaps for dimensionality reduction and data
  representation}.
\newblock {\em Neural Computation}, 15(6):1373--1396, 2003.

\bibitem{bergstra2012random}
James Bergstra and Yoshua Bengio.
\newblock {Random search for hyper-parameter optimization}.
\newblock {\em Journal of Machine Learning Research}, 13(Feb):281--305, 2012.

\bibitem{berry2016local}
Tyrus Berry and Timothy Sauer.
\newblock {Local kernels and the geometric structure of data}.
\newblock {\em Applied and Computational Harmonic Analysis}, 40(3):439--469,
  2016.

\bibitem{cardoso2007}
Ana Cardoso-Cachopo and Arlindo~L Oliveira.
\newblock {Semi-supervised single-label text categorization using
  centroid-based classifiers}.
\newblock In {\em Proceedings of the 2007 ACM symposium on Applied computing},
  pages 844--851. ACM, 2007.

\bibitem{coifman2006diffusion}
Ronald~R Coifman and St{\'e}phane Lafon.
\newblock {Diffusion maps}.
\newblock {\em Applied and computational harmonic analysis}, 21(1):5--30, 2006.

\bibitem{de2010optimized}
Kris De~Brabanter, Jos De~Brabanter, Johan~AK Suykens, and Bart De~Moor.
\newblock {Optimized fixed-size kernel models for large data sets}.
\newblock {\em Computational Statistics \& Data Analysis}, 54(6):1484--1504,
  2010.

\bibitem{drineas2005nystrom}
Petros Drineas and Michael~W Mahoney.
\newblock {On the Nystr{\"o}m method for approximating a Gram matrix for
  improved kernel-based learning}.
\newblock {\em Journal of Machine Learning Research}, 6:2153--2175, Dec 2005.

\bibitem{drineas2018lectures}
Petros Drineas and Michael~W Mahoney.
\newblock {Lectures on randomized numerical linear algebra}.
\newblock {\em The Mathematics of Data}, 25:1, 2018.

\bibitem{BurkeKendall2014}
Graham~Kendall Edmund K.~Burke.
\newblock {\em {Search methodologies: Introductory tutorials in optimization
  and decision support techniques}}.
\newblock Springer US, New York, NY, USA, 2014.

\bibitem{endres2003new}
Dominik~Maria Endres and Johannes~E Schindelin.
\newblock {A new metric for probability distributions}.
\newblock {\em IEEE Transactions on Information theory}, 2003.

\bibitem{fayed2007}
Hatem~A Fayed, Sherif~R Hashem, and Amir~F Atiya.
\newblock {Self-generating prototypes for pattern classification}.
\newblock {\em Pattern Recognition}, 40(5):1498--1509, 2007.

\bibitem{genton2001classes}
Marc~G Genton.
\newblock {Classes of kernels for machine learning: A statistics perspective}.
\newblock {\em Journal of machine learning research}, 2(Dec):299--312, 2001.

\bibitem{gonzalez1985}
Teofilo~F Gonzalez.
\newblock {Clustering to minimize the maximum intercluster distance}.
\newblock {\em Theoretical Computer Science}, 38:293--306, 1985.

\bibitem{guo2014feature}
Jianhui Guo, Ping Yi, Ruili Wang, Qiaolin Ye, and Chunxia Zhao.
\newblock {Feature selection for least squares projection twin support vector
  machine}.
\newblock {\em Neurocomputing}, 144:174--183, 2014.

\bibitem{he2018kernel}
Li~He and Hong Zhang.
\newblock {Kernel K-means sampling for Nystr{\"o}m approximation}.
\newblock {\em IEEE Transactions on Image Processing}, 27(5):2108--2120, 2018.

\bibitem{hochbaum1985}
Dorit~S Hochbaum and David~B Shmoys.
\newblock {A best possible heuristic for the k-center problem}.
\newblock {\em Mathematics of Pperations Research}, 10(2):180--184, 1985.

\bibitem{hofmann2008kernel}
Thomas Hofmann, Bernhard Sch{\"o}lkopf, and Alexander~J Smola.
\newblock {Kernel methods in machine learning}.
\newblock {\em The annals of statistics}, pages 1171--1220, 2008.

\bibitem{homrighausen2016nystrom}
Darren Homrighausen and Daniel~J McDonald.
\newblock {On the Nystr{\"o}m and column-sampling methods for the approximate
  principal components analysis of large datasets}.
\newblock {\em Journal of Computational and Graphical Statistics},
  25(2):344--362, 2016.

\bibitem{Joachims/99a}
T.~Joachims.
\newblock {Making large-scale SVM learning practical}.
\newblock In B.~Sch{\"o}lkopf, C.~Burges, and A.~Smola, editors, {\em Advances
  in Kernel Methods - Support Vector Learning}, chapter~11, pages 169--184. MIT
  Press, Cambridge, MA, 1999.

\bibitem{john2010elements}
ZQ~John~Lu.
\newblock {The elements of statistical learning: data mining, inference, and
  prediction}.
\newblock {\em Journal of the Royal Statistical Society: Series A (Statistics
  in Society)}, 173(3):693--694, 2010.

\bibitem{kumar2012sampling}
Sanjiv Kumar, Mehryar Mohri, and Ameet Talwalkar.
\newblock {Sampling methods for the Nystr{\"o}m method}.
\newblock {\em Journal of Machine Learning Research}, 13(Apr):981--1006, 2012.

\bibitem{li2016kernel}
Ximing Li, Jihong Ouyang, and Xiaotang Zhou.
\newblock {A kernel-based centroid classifier using hypothesis margin}.
\newblock {\em Journal of Experimental \& Theoretical Artificial Intelligence},
  28(6):955--969, 2016.

\bibitem{liu2017new}
Chuan Liu, Wenyong Wang, Guanghui Tu, Yu~Xiang, Siyang Wang, and Fengmao Lv.
\newblock {A new centroid-based classification model for text categorization}.
\newblock {\em Knowledge-Based Systems}, 136:15--26, 2017.

\bibitem{liu2016feature}
Jie Liu and Enrico Zio.
\newblock {Feature vector regression with efficient hyperparameters tuning and
  geometric interpretation}.
\newblock {\em Neurocomputing}, 218:411--422, 2016.

\bibitem{liu2004improving}
Qingshan Liu, Hanqing Lu, and Songde Ma.
\newblock {Improving kernel Fisher discriminant analysis for face recognition}.
\newblock {\em IEEE transactions on circuits and systems for video technology},
  14(1):42--49, 2004.

\bibitem{tass2017}
Eugenio Martinez-Camara, Manuel~C. Diaz-Galiano, M.~Angel Garcia-Cumbreras,
  Manuel Garcia-Vega, and Julio Villena-Roman.
\newblock {Overview of TASS 2017}.
\newblock In {\em TASS 2017: Workshop on Semantic Analysis at SEPLN}, pages
  13--21, 2017.

\bibitem{mika1999fisher}
Sebastian Mika, Gunnar Ratsch, Jason Weston, Bernhard Scholkopf, and
  Klaus-Robert Mullers.
\newblock {Fisher discriminant analysis with kernels}.
\newblock In {\em Neural networks for signal processing IX, 1999. Proceedings
  of the 1999 IEEE signal processing society workshop.}, pages 41--48. Ieee,
  1999.

\bibitem{semeval2017}
Sabino Miranda-Jim{\'e}nez, Mario Graff, Eric~Sadit Tellez, and Daniela
  Moctezuma.
\newblock {INGEOTEC at SemEval 2017 Task 4: A B4MSA ensemble based on genetic
  programming for Twitter sentiment analysis}.
\newblock 2017.

\bibitem{muandet2017kernel}
Krikamol Muandet, Kenji Fukumizu, Bharath Sriperumbudur, Bernhard
  Sch{\"o}lkopf, et~al.
\newblock {Kernel mean embedding of distributions: A review and beyond}.
\newblock {\em Foundations and Trends in Machine Learning}, 10(1-2):1--141,
  2017.

\bibitem{murphy2012machine}
Kevin~P Murphy.
\newblock {\em {Machine learning: A probabilistic perspective}}.
\newblock MIT press, 2012.

\bibitem{musco2017recursive}
Cameron Musco and Christopher Musco.
\newblock {Recursive sampling for the Nystr{\"o}m method}.
\newblock In {\em Advances in Neural Information Processing Systems}, pages
  3833--3845, 2017.

\bibitem{sklearn}
F.~Pedregosa, G.~Varoquaux, A.~Gramfort, V.~Michel, B.~Thirion, O.~Grisel,
  M.~Blondel, P.~Prettenhofer, R.~Weiss, V.~Dubourg, J.~Vanderplas, A.~Passos,
  D.~Cournapeau, M.~Brucher, M.~Perrot, and E.~Duchesnay.
\newblock {Scikit-learn: Machine Learning in Python}.
\newblock {\em Journal of Machine Learning Research}, 12:2825--2830, 2011.

\bibitem{raicharoen2005}
Thanapant Raicharoen and Chidchanok Lursinsap.
\newblock {A divide-and-conquer approach to the pairwise opposite class-nearest
  neighbor (POC-NN) algorithm}.
\newblock {\em Pattern recognition letters}, 26(10):1554--1567, 2005.

\bibitem{sanchez2004}
Jos{\'e}~Salvador S{\'a}nchez.
\newblock {High training set size reduction by space partitioning and prototype
  abstraction}.
\newblock {\em Pattern Recognition}, 37(7):1561--1564, 2004.

\bibitem{scardapane2019kafnets}
Simone Scardapane, Steven Van~Vaerenbergh, Simone Totaro, and Aurelio Uncini.
\newblock {Kafnets: Kernel-based non-parametric activation functions for neural
  networks}.
\newblock {\em Neural Networks}, 110:19--32, 2019.

\bibitem{sheikhpour2017kernelized}
Razieh Sheikhpour, Mehdi~Agha Sarram, Mohammad Ali~Zare Chahooki, and Robab
  Sheikhpour.
\newblock {A kernelized non-parametric classifier based on feature ranking in
  anisotropic Gaussian kernel}.
\newblock {\em Neurocomputing}, 267:545--555, 2017.

\bibitem{suykens1999least}
Johan~AK Suykens and Joos Vandewalle.
\newblock {Least squares support vector machine classifiers}.
\newblock {\em Neural Processing Letters}, 9(3):293--300, 1999.

\bibitem{tang2014feature}
Jiliang Tang, Salem Alelyani, and Huan Liu.
\newblock {Feature selection for classification: A review}.
\newblock {\em Data classification: Algorithms and applications}, page~37,
  2014.

\bibitem{triguero2012}
Isaac Triguero, Joaqu{\'\i}n Derrac, Salvador Garcia, and Francisco Herrera.
\newblock {A taxonomy and experimental study on prototype generation for
  nearest neighbor classification}.
\newblock {\em IEEE Transactions on Systems, Man, and Cybernetics, Part C
  (Applications and Reviews)}, 42(1):86--100, 2012.

\bibitem{vapnik2013nature}
Vladimir Vapnik.
\newblock {\em {The nature of statistical learning theory}}.
\newblock Springer science \& business media, 2013.

\bibitem{wang2019scalable}
Shusen Wang, Alex Gittens, and Michael~W Mahoney.
\newblock {Scalable kernel K-means clustering with Nystr{\"o}m approximation:
  relative-error bounds}.
\newblock {\em The Journal of Machine Learning Research}, 20(1):431--479, 2019.

\bibitem{williams2001using}
Christopher~KI Williams and Matthias Seeger.
\newblock {Using the Nystr{\"o}m method to speed up kernel machines}.
\newblock In {\em Advances in neural information processing systems}, pages
  682--688, 2001.

\bibitem{yeo2000new}
In-Kwon Yeo and Richard~A Johnson.
\newblock {A new family of power transformations to improve normality or
  symmetry}.
\newblock {\em Biometrika}, 87(4):954--959, 2000.

\bibitem{zhang2009density}
Kai Zhang and James~T Kwok.
\newblock {Density-weighted Nystr{\"o}m method for computing large kernel
  eigensystems}.
\newblock {\em Neural Computation}, 21(1):121--146, 2009.

\bibitem{zhang2010clustered}
Kai Zhang and James~T Kwok.
\newblock {Clustered Nystr{\"o}m method for large scale manifold learning and
  dimension reduction}.
\newblock {\em IEEE Transactions on Neural Networks}, 21(10):1576--1587, 2010.

\bibitem{zhuang2011family}
Jinfeng Zhuang, Ivor~W Tsang, and Steven~CH Hoi.
\newblock {A family of simple non-parametric kernel learning algorithms}.
\newblock {\em Journal of Machine Learning Research}, 12(Apr):1313--1347, 2011.

\end{thebibliography}
\end{document}